\pdfoutput=1
\documentclass[11pt]{article}

\usepackage[]{acl}

\usepackage[dvipsnames]{xcolor}
\usepackage{times}
\usepackage{latexsym}
\usepackage{makecell}

\usepackage[LFE,LAE,T1]{fontenc}  
\usepackage[utf8]{inputenc}       
\usepackage[arabic,english]{babel} 

\usepackage[T1]{fontenc}

\usepackage{microtype}

\usepackage{inconsolata}
\usepackage{booktabs}

\usepackage{graphicx}
\graphicspath{ {./img/} }
\newcommand{\benchmark}{Arab Voices}

\newcommand{\peter}[1]{\noindent{\textcolor{ForestGreen}{\{{$_{Peter}$} \em   #1\}}}}

\usepackage{comment}
\usepackage{booktabs}
\usepackage{threeparttable}
\usepackage{multirow}
\usepackage{soul}
\usepackage{arydshln}
\usepackage{colortbl}
\usepackage{amssymb}
   \usepackage{cuted}
\usepackage{enumitem}

\usepackage{appendix}
\usepackage{amsmath}
\usepackage{hyperref}
\usepackage[normalem]{ulem} 

\usepackage{caption}
\definecolor{lowscore}{RGB}{220, 50, 50}  
\definecolor{highscore}{RGB}{0, 100, 0}   

\newcommand{\low}[1]{\textcolor{lowscore}{#1}}
\newcommand{\high}[1]{\textcolor{highscore}{\textbf{#1}}}

\usepackage{array} 
\newcolumntype{H}{>{\setbox0=\hbox\bgroup}c<{\egroup}@{}}

\title{\raisebox{-0.29\height}{\includegraphics[scale=0.025]{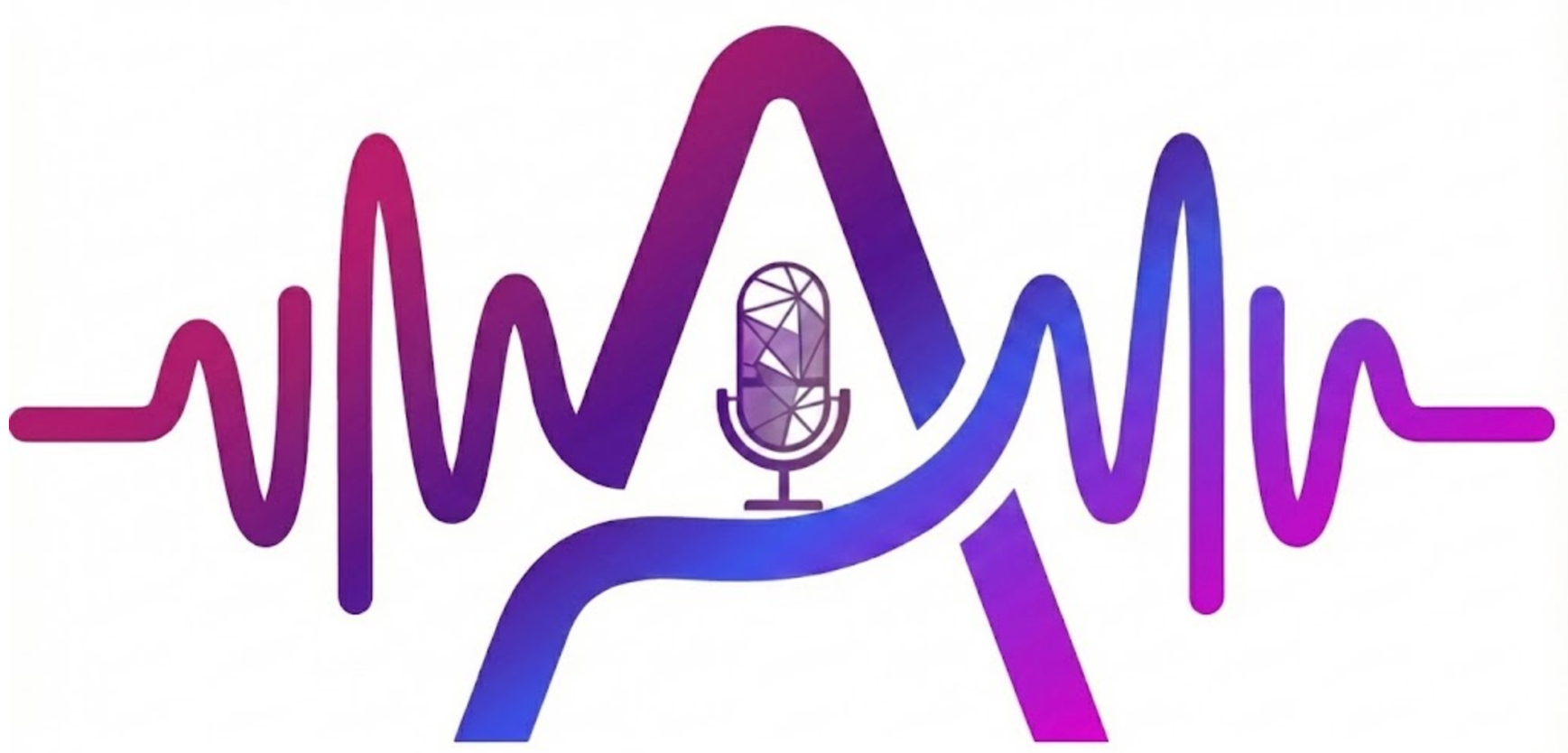}}\benchmark: Mapping Standard and Dialectal Arabic Speech Technology}

\author{\normalsize Peter Sullivan$^{\xi}$ ~~ AbdelRahim Elmadany$^{\xi}$  ~~ Alcides Alcoba Inciarte$^{\xi}$ ~~ Muhammad Abdul-Mageed$^{\xi,\lambda}$\\
\normalsize $^{\xi}$The University of British Columbia ~~ $^{\lambda}$Canada Research Chair in NLP and ML \\%
  \texttt{\normalsize \{prsull@student.,a.elmadany@,muhammad.mageed@,alcobaaj@mail.\}ubc.ca}}

\begin{document}
\maketitle

\begin{strip}
  \centering
  \includegraphics[width=\textwidth]{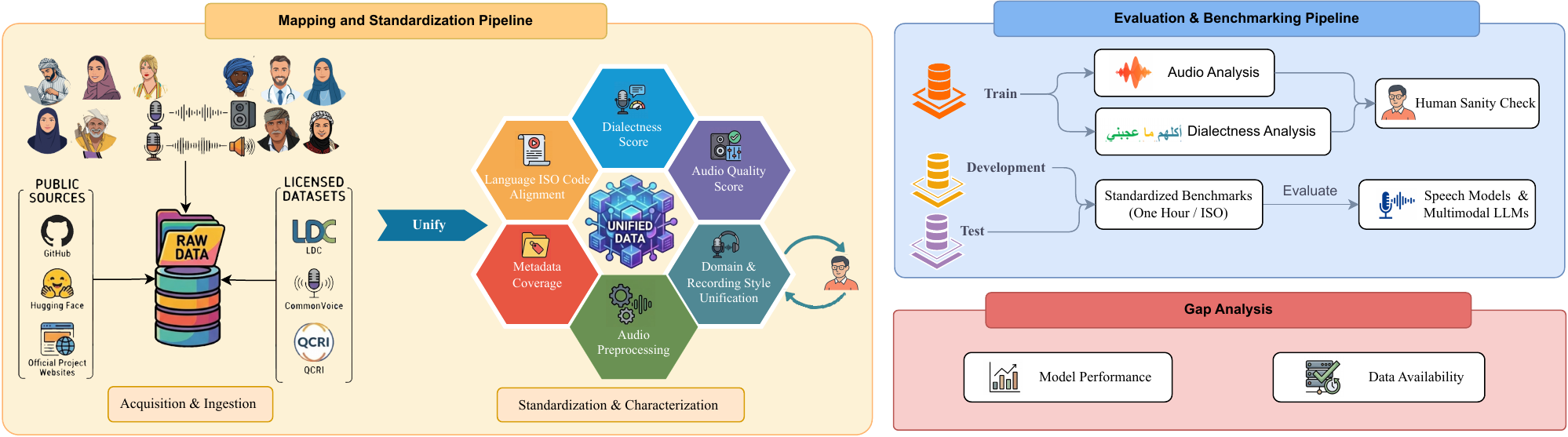}
  \vspace{-1.5em}
  \captionof{figure}{Mapping framework for standard and dialectal Arabic speech technology.
  We unify public and licensed datasets through a standardization pipeline, and then profile their demographic and quality characteristics, validating results through human-in-loop review. 
  The unified data are then used to 
  benchmark 
  speech models and multimodal LLMs; the results of which 
  alongside our dataset profiling, helps informs 
  future data collection and modeling efforts.}
  \label{fig:participants_distri}
\end{strip}


\begin{abstract}
Dialectal Arabic (DA) speech data vary widely in domain coverage, dialect labeling practices, and recording conditions, complicating cross-dataset comparison and model evaluation. To characterize this landscape, we conduct a computational analysis of linguistic ``dialectness'' alongside objective proxies of audio quality on the training splits of widely used DA corpora. We find substantial heterogeneity both in acoustic conditions and in the strength and consistency of dialectal signals across datasets, underscoring the need for standardized characterization beyond coarse labels. To reduce fragmentation and support reproducible evaluation, we introduce \benchmark, a standardized framework for DA ASR. \benchmark~provides unified access to 31 datasets spanning 14 dialects, with harmonized metadata and evaluation utilities. We further benchmark a range of recent ASR systems, establishing strong baselines for modern DA ASR.

\end{abstract}

\section{Introduction}
The landscape of Dialectal Arabic (DA) speech processing remains fragmented. While Modern Standard Arabic (MSA) has relatively robust computational support, the diverse spoken varieties of Arabic, the primary medium of daily communication for $\sim$450 million native of speakers, remain comparatively underserved. Recent surveys of DA speech datasets \cite{alqadasi2025arabic} highlight recurring bottlenecks, including limited open-access standardization, uneven coverage of underrepresented varieties, inconsistent sub-dialect labeling, and insufficient documentation of data quality and collection conditions. A central challenge is the lack of standardized metadata that would allow practitioners to align training data with target applications. dialect labels, for instance, may be defined by geopolitical boundaries \cite{talafha2024casablanca,ali2019mgb}, coarse regional groupings \cite{ali2017speech}, or ISO-style codes (e.g., \textit{apc}, \textit{ary}) \cite{LDC2006S43_gulf_speech,LDC2006S45_iraqi_speech}, which hinders principled pooling and selection of data for low-resource varieties. At the same time, pooling is not universally beneficial: variation in mutual intelligibility across the Arabic continuum can make indiscriminate mixing less effective than targeted cross-dialect transfer \cite{talafha2024casablanca}. Richer, measurable characterization of candidate datasets can therefore help guide dataset selection and model design.

While existing surveys make these gaps visible, addressing them in practice requires a unified and reproducible framework. To this end, we introduce \benchmark, a standardized ecosystem for assembling heterogeneous DA corpora into a consistent format, with harmonized metadata that supports comparison across datasets. In addition to reconciling labeling inconsistencies, our framework supports dialect organization beyond country-level tags, enabling more granular, linguistically informed representations where available. We also revisit what ``quality'' should mean for DA speech resources. In this setting, acoustic fidelity and linguistic authenticity do not always coincide: studio recordings can offer clean audio but may under-represent spontaneous phenomena common in everyday speech, such as code-switching, disfluencies, and prosodic variability. Moreover, documentation of sociolinguistic variables (e.g., urban vs.\ rural speech, speaker demographics) is often limited or absent. We therefore complement metadata standardization with automated dataset characterization, quantifying both acoustic conditions and linguistic ``dialectness'' to support principled dataset selection and more robust downstream modeling.

In this paper, we build a foundation for systematically analyzing and benchmarking DA speech technology. Our contributions are: (I) \textbf{Standardized mapping.} We provide a uniform framework for aggregating and mapping heterogeneous DA datasets into a common format, including harmonized metadata and resolved dialect-label inconsistencies. (II) \textbf{Automated enrichment.} We use automated methods to characterize each dataset along two axes, i.e., linguistic ``dialectness'' and objective audio-quality proxies, and analyze variability across training splits. (III) \textbf{Multi-dialect benchmark.} We introduce an ASR benchmark spanning 31 datasets and 14 dialects, including varieties previously identified as underrepresented. (IV) \textbf{Broad evaluation.} We evaluate a diverse set of recent open-weight ASR systems, ranging from speech-centric architectures to audio-capable multimodal foundation models, establishing baselines for modern DA ASR.


\section{A Framework for Real-World Variability in DA Speech Data}
\label{sec:motivation}

DA speech technologies are often trained and evaluated on datasets that differ substantially in linguistic coverage and recording conditions, yet these differences are not always documented in a way that supports reproducible comparison. In an ideal setting, training data would reflect the variability of real-world speech along both linguistic and acoustic axes. We therefore adapt the dataset documentation perspective of \citet{bender2018data} to DA speech corpora, focusing on dimensions that directly shape observable variability in audio and transcripts, and adding concrete descriptors of recording quality.\footnote{Broader documentation dimensions such as curation rationale, provenance, and annotator demographics are important, but we treat them as complementary to the variability-focused axes studied here and leave their systematic treatment to future work.} Therefore, we use this framework to guide what we standardize and what we quantify across DA datasets.

\paragraph{Linguistic variety.}
A primary source of variability is \textit{regional language variation}: communities may differ in lexical choice, phonology, prosody, and morphosyntax. For practical data organization, we distinguish \textbf{dialectal variation} from \textbf{regional accent}, where the latter is more plausibly characterized by primarily phonetic differences within a shared dialectal lexicon/grammar \citep[p.~43]{wardhaugh2021introduction}. This distinction matters for dataset aggregation, since labels derived from geography or ISO codes can conflate lexical/grammatical differences with pronunciation differences, affecting both pooling decisions and the interpretation of model errors. Therefore, we harmonize dialect metadata and explicitly track dialect/region labels in a consistent schema.

\paragraph{Speech situation and speaker factors.}
Beyond region, DA corpora vary with \textit{speech style} and social situation, which shape register and expressive content \citealp[p.~69,74--77]{Eckert_2016,irvine2002style}. Concretely, we highlight \textbf{register} (language conditioned by activity or setting) \citealp[p.~48]{wardhaugh2021introduction}; \textbf{affect} (emotion expressed through speech) \citealp[p.~80]{Eckert_2016}; and speaker-related factors such as age, gender, education, and socioeconomic background, which can correlate with systematic linguistic differences \citealp[p.~75--76,80]{Eckert_2016,wardhaugh2021introduction,lee2025language}. In many released DA datasets, however, these sociolinguistic attributes are either missing or inconsistently reported, limiting our ability to stratify evaluation by speaker context. Therefore, we treat these factors as desiderata in our documentation schema and focus our automated characterization on measurable proxies available at scale.

\paragraph{Recording and channel conditions.}
A third major axis is the recording process itself. Device and \textbf{channel} characteristics can affect model behavior, and environmental conditions introduce reverberation and background noise that vary with room acoustics and microphone placement \citep{khokhlov24_interspeech,ryu25b_interspeech}. Dataset-level technical choices such as \textbf{sampling rate} and \textbf{file format} can also influence some speech tasks and evaluation comparability \citep{ferrofilho25_interspeech}. Because these factors often go under-reported, we complement metadata with objective \textit{audio-quality proxies} computed directly from the training audio to enable consistent, cross-corpus comparison. 

While this framework does not exhaustively capture all possible sources of variability, it provides a task-oriented lens for documenting and comparing DA speech resources and for identifying systematic gaps. Guided by it, we (i) standardize heterogeneous DA datasets into a unified format with harmonized dialect metadata and (ii) automatically characterize each dataset along two measurable axes (i.e., linguistic ``dialectness'' and acoustic conditions) to support principled dataset selection and robust benchmarking. Therefore, we next turn to mapping and analyzing the current landscape of spoken Arabic data.

\section{Literature Review}
\label{sec:lit_review}

\paragraph{`Real-World' Performance}
Speech recognition models often exhibit a gap between performance on curated benchmarks and performance under real-world, out-of-domain conditions \cite{likhomanenko2020rethinking,radford2023robust,sullivan23_interspeech}. Although true deployment conditions are difficult to anticipate, cross-corpus evaluation over diverse datasets can approximate real-world variability and provide a more informative stress test of robustness \cite{likhomanenko2020rethinking}. Common approaches for narrowing this gap include large-scale supervised pretraining \cite{radford2023robust}, training objectives designed to reduce reliance on dataset-specific text style \cite{pratap2024scaling}, and augmentation with noise and reverberation \cite{likhomanenko2020rethinking}. In Arabic ASR, however, evaluation is still frequently reported on a single (often in-domain) benchmark \cite{ali2016mgb,ali2017speech,ali2019mgb,talafha2024casablanca}, and comparatively fewer studies report systematic out-of-domain results, which have highlighted substantial remaining challenges for dialectal settings \cite{talafha2024casablanca}. Therefore, our work emphasizes cross-dataset benchmarking to better reflect the variability encountered in Dialectal Arabic.

\paragraph{Spoken Dialectal Arabic}
Several systematic reviews have surveyed DA datasets \cite{alqadasi2025arabic}, MSA datasets \cite{alqadasi2023modern}, and Arabic speech technologies including dialect identification \cite{elnagar2021systematic}, ASR \cite{alsayadi2022deep}, and TTS \cite{chemnad2023advancements}. Collectively, these surveys summarize trends in dataset creation and intended use \cite{alqadasi2023modern,alqadasi2025arabic}, reported demographic categories \cite{alqadasi2023modern,alqadasi2025arabic}, recording environments \cite{alqadasi2023modern,alqadasi2025arabic}, and the coverage of Arabic varieties \cite{alqadasi2025arabic,chemnad2023advancements,elnagar2021systematic,alsayadi2022deep}. However, such reviews typically rely on what is reported in papers and do not directly audit dataset properties; for example, while \citet{alqadasi2023modern,alqadasi2025arabic} tabulate which demographic attributes are mentioned, they do not quantify the distribution of utterance counts or durations across demographic groups.

Taxonomies of Arabic dialects are also inconsistent across prior work. Some studies adopt coarse regional groupings (e.g., Gulf, North African) \cite{ali2017speech,abdullah25_interspeech}, others use country-level labels \cite{ali2019mgb,talafha2024casablanca,elnagar2021systematic,alsayadi2022deep,alqadasi2025arabic}, and still others propose more fine-grained schemas \cite{alharbi2024sada,omnilingual2025omnilingual}. 
We provide harmonized labeling and complementary, automated characterization to support more consistent comparison across DA datasets. We cover audio data quality and Speech Processing Methods in Appendix \S\ref{appdx_sec:lit_review}.
\section{Standardization and Mapping}
\label{sec:mapping}
\begin{table*}[!h]
\centering
\resizebox{0.9\textwidth}{!}{%
\begin{tabular}{cllrclll}
\toprule
& \textbf{Dataset} & \textbf{Shorthand} & \textbf{Duration(h)} & \textbf{Variety} & \textbf{Country} & \textbf{Rec. Style}  \\ \midrule 

\multirow{29}{*}{\rotatebox{90}{\textbf{\colorbox{blue!15}{ASR}}}} & ArVoice~\cite{toyin2025arvoicemultispeakerdatasetarabic} & ARVOI & 6 8.3 & arb & N/A & Read (\textit{c}) \\
& ArzEn~\cite{hamed2020arzen} & ARZEN & 11.4 & arz & Egypt & Conv.(\textit{d}) \\
& CALLHOME~\cite{LDC97S45_callhome_egypt} & CALLH & 17  & arz  & Egypt & Conv. \\
& Cassablanca~\cite{talafha2024casablanca} & CASSA & 13.8 & \makecell{afb, apc, arq, ary, arz, mey}  & Various & YouTube \\
& Common Voice 22 (\textit{g})~\cite{ardila2019common} & CV22 & 58.1 & arb & Global & Read\\
& Egyptian Conv~\cite{magicdata_no_date_egyconv} & EGCON & 2.3 & arz, acq & Egypt & Conv. \\
& FLEURS~\cite{conneau2023fleurs} & FLEUR & 8.2  & arb & Egypt & Read  \\
& Iraqi Telephone~\cite{LDC2006S45_iraqi_speech} & IRQTL & 40.2 & acm, ayp & Iraq & Conv. \\
& IWSLT~\cite{agostinelli-etal-2025-findings} & IWSLT & 2 & aeb, apc & Various & Various \\
& GALE(\textit{a})~\cite{LDC2013S07_GALE_p2_conv_1} & GALE & 972.9 & arb & Various &  Broadcast\\
& Gulf Telephone~\cite{LDC2006S43_gulf_speech} & GULTL & 97.9 & afb & Various & Conv. \\

& L2 KSU~\cite{LDC2024S11_l2_ksu} & L2KSU & 7.7 & arb & KSA & Read  \\
& LINTO~\cite{naouaralinto} & LINTO &  1.4 &  aeb & Tunisia & Various \\
& MASC~\cite{al2023masc} & MASC & 518.2 & arb & Various & YouTube \\
& MediaSpeech~\cite{mediaspeech2021} & MS & 10 & arb & Various & Broadcast \\
& MGB2~\cite{ali2016mgb} & MGB2 & 1,147.5 & arb  & Various & Broadcast  \\
& MGB3~\cite{ali2017speech} & MGB3 & 12.2  & arz  & Egypt & YouTube \\
& MGB5~\cite{ali2019mgb} & MGB5 & 54  & ary  & Morocco & YouTube \\
& QASR~\cite{mubarak2021qasr} & QASR & 1,914.5 & arb & Various & YouTube \\
& Quran Speech~\cite{slr132} & QURAN & 795.6 & arb(\textit{b}) & Various & Recitation \\
& SADA2022~\cite{alharbi2024sada} &SADA & 437.6 & acw, ars & KSA &  Broadcast\\
& SCC22~\cite{SCC2025} & SCC22 & 4.5 & UNK(\textit{e})  & KSA & Podcast (\textit{d}) \\
& TARIC-SLU~\cite{mdhaffar2024taric} & TARIC & 0.9 & aeb & Tunisia & Various \\
& Tunisian MSA~\cite{slr46} & TUMSA & 8.7 & arb & Tunisia & Read \\
& Yemeni Conv~\cite{magicdata_no_date_yemconv} & YEMTL & 4.8 & ayn, acq & Yemen & Conv. \\
& ZAEBUC~\cite{habash2022zaebuc} & ZAEBC & 0.4 & arb, arz, afb  & UAE & Discussion (\textit{d}) \\ \midrule

\multirow{3}{*}{\rotatebox{90}{\textbf{\colorbox{green!15}{TTS}}}} & Arabic-Diacritized-TTS~\cite{NourhannADTTS} & ADTTS &  39.2 & arb  & N/A & Read (\textit{c})\\
& ClArTTS~\cite{kulkarni2023clartts} & CATTS & 11.2 & arb(\textit{b})  & Various & Read \\
& Iraqi TTS~\cite{kharrufa_2024_11170567} & IRTTS & 5.0 & acm & Iraq & Read \\ \midrule
\multirow{2}{*}{\textbf{\colorbox{red!15}{EMO}}} & KSUEmotion~\cite{LDC2024S11_l2_ksu} & KSUE & 5.2 & arb & KSA & Read  \\
& KEDAS~\cite{LDC2023S10_kasdi} & KEDAS & 2.1 & arb & Algeria & Read  \\
\bottomrule

\end{tabular}%
}
\caption{Transcribed Audio datasets for Arabic.  We provide total hours of the filtered datasets (removing very short utterances and non-Arabic), as well as the primary varieties of Arabic spoken. For recording styles, we identify the the major kinds of recording when possible, and highlight the ambiguity of this terms with regards to recordings sourced from YouTube. We abbreviate conversational as `Conv.' We note recording style as `Various' when it may consist of distinct sources. (\textit{a}) We include Parts 2, 3, and 4 of GALE. (\textit{b}) Classical Arabic, (\textit{c}) Partially synthetic, (\textit{d}) Codeswitching-focused, (\textit{e}) Underspecified dialect, (\textit{g}) reaching out to CommonVoice directly to obtain.}
\label{tab:arabic_data}
\vspace{-1em}
\end{table*}


This section describes our end-to-end pipeline for (i) curating DA speech datasets, (ii) standardizing audio and text into a common representation, (iii) harmonizing dialect and domain metadata, and (iv) constructing a benchmark with a standardized training/adaptation protocol.

\subsection{Standardization}

\paragraph{Dataset curation.}
We identify candidate datasets through two complementary routes: (1) openly available repositories (e.g., OpenSLR and the Hugging Face Datasets hub), and (2) scholarly search for papers that introduce or document speech corpora for specific Arabic varieties, followed by retrieval of their corresponding releases when available.
We fix a cutoff date of September~2025.
Table~\ref{tab:arabic_data} summarizes the resulting collection, and \S\ref{appdx_sec:mapping} provides dataset-specific notes and provenance.

\paragraph{Audio and text preprocessing.}
To enable consistent downstream processing and evaluation, we convert all audio to a standardized representation: mono, 16~kHz sampling rate, and 16-bit PCM encoding.
For corpora released as two-channel conversational audio with per-channel speakers (e.g., \citealp{LDC2006S43_gulf_speech}), we split channels and align each channel to its corresponding speaker transcript when such alignment is provided.
For other multi-channel recordings without speaker-channel semantics, we downmix to mono by averaging channels. We retain the original transcript as a raw field and create a standardized text field for analysis and scoring.
We convert Buckwalter transliteration to Arabic script when present, and we preserve diacritics and author-provided annotations in the raw field.
For the standardized field, we remove segments that are entirely Latin-script (e.g., metadata-only lines) and retain mixed-script utterances.
We apply light normalization intended to improve comparability across corpora while minimizing semantic changes (details in \S\ref{sec:analysismethods} and \S\ref{appdx_sec:mapping}).

Each utterance is represented with a common \textbf{schema} that includes (when available): a unique utterance ID, audio path, duration, raw transcript, standardized transcript, dataset/source identifiers, speaker ID, recording metadata, domain labels, and dialect labels.
This standardized representation is used both for dataset characterization and for benchmark construction.

\subsection{Mapping and Metadata Harmonization}

\paragraph{Dialect alignment.}

 To harmonize dialect labels across heterogeneous annotation practices, we align dialects as best as possible using a combination of ISO 639-3 language codes plus country-region codes. This combination provides a precise and compact description of the type of language being spoken (e.g. afb\_ARE-AZ: Gulf/Khaleeji Arabic as spoken by someone in Abu Dhabi).
For many datasets, the ISO 639-3 code can be inferred from country and city information, however, this provides problems for dialects from countries with multiple major dialects (e.g. Saudi Arabia), which often remain ambiguous without speaker information.
In cases where the location within a country is known we use Ethnologue's language maps~\cite{Eberhard2025ethnologue} to better align to ISO language description. 

\paragraph{Domain normalization.}
Domain labels are inconsistently named across datasets (e.g., overlapping or near-synonymous descriptions).
Across the 61 distinct domain strings observed in metadata, we manually normalize synonymous and closely related labels (e.g., \textit{places to go} and \textit{travel}) into 11 broad themes used for reporting and stratification.
We retain both the original domain string and the normalized theme to support reproducibility.

\paragraph{Label sanity check.}
To assess gross inconsistencies in mapped labels, we conduct an initial manual review of a stratified sample of utterances ($n=83$), focusing on dialect and domain tags where ambiguity or label drift is most likely.
Findings from this check inform minor corrections to the mapping rules and highlight datasets requiring conservative treatment in the benchmark (e.g., exclusion due to unresolved ambiguity).

\subsection{Automated Characterization and Analysis}
\label{sec:analysismethods}

Guided by the variability dimensions in Table~\ref{tab:ideal_framework}, we first scan each dataset's \emph{training split} for available metadata.\footnote{This analysis is limited to 28 datasets with accessible training material; the remaining datasets are evaluation-only.}
We then quantify two measurable axes at scale: (i) linguistic ``dialectness'' and (ii) acoustic/recording conditions, which together provide a more informative characterization than coarse labels alone.
We report aggregate statistics per dataset and per dialect grouping, and we analyze how these signals vary across corpora.

\paragraph{Text analysis (dialectness).}
We apply two complementary approaches.
First, we use an in-house binary MSA-DA classifier to estimate whether an utterance exhibits primarily MSA-like vs.\ dialectal lexical/orthographic patterns.
Second, we apply Arabic Level of Dialectness (ALDi) \citep{keleg2023aldi}, which provides a graded estimate of dialectal intensity.
We aggregate these utterance-level outputs to dataset-level distributions, enabling comparisons across corpora and supporting detection of potential label mismatches (e.g., corpora labeled as DA but exhibiting strongly MSA-like transcripts).

\paragraph{Audio-quality analysis.}
To characterize recording conditions without matched clean references, we use TorchAudio-SQUIM \citep{kumar2023torchaudio} to \emph{predict} no-reference proxies of common speech quality/intelligibility measures.
Specifically, we compute predicted PESQ \citep{rix2001perceptual}, predicted STOI \citep{taal2011algorithm}, predicted SI-SDR \citep{le2019sdr}, and a predicted no-reference MOS (NMR-MOS) \citep{manocha22c_interspeech}.
We treat these as comparative proxies rather than ground-truth scores and analyze their distributions across datasets and dialect groupings.
Figure~\ref{fig:audio_landscape} summarizes key trends.

\subsection{Human Sanity Check}
\label{sec:human_validation}

Because automated proxies can fail in systematic ways, we conduct a lightweight human sanity check to contextualize the signals used in our analysis.
We sample approximately 200 utterances (stratified across datasets and dialect labels where possible) for manual inspection by a native Arabic speaker.
For dialectness, the annotator assigns ALDi bins following \citet{keleg2023aldi}.
For audio, the annotator provides a 1-5 MOS-style rating anchored using examples from the VoiceMOS~2022 challenge.\footnote{\url{https://zenodo.org/records/6572573}}
We use this study to verify that (i) extreme-score cases correspond to perceptible differences and (ii) the automated measures are directionally aligned with human judgments, while treating it as a sanity check rather than a full-scale validation.

\subsection{Benchmarking and Split Release}
\label{sec:benchmark}

To build our final benchmark we identify the dev and test splits of each dataset where available and ensure these materials retain their original split designation. For dialects lacking datasets with canonical splits, if there is enough data we sample one hour for test and dev, retaining the rest for train. Where we do not have enough for this, we evenly split data across train, dev, and test.
For datasets with ambiguous (more than one possible ISO match e.g. `Maghrebi') we exclude these from the final benchmark. 
Subsets corresponding to localities are also provided, for the samples where this information is available.

In total we arrive with a benchmark covering 14 of the ISO 639-3 labels,  with each dialect provide with an adaptation split of five hours, one hour development, and a one hour test split. These splits pull from multiple datasets where possible (see Table \ref{tab:benchmark} for detailed breakdown of composition), in order to better reflect performance on diverse domains and conditions. Due to license and copyright restrictions, we do not provide the dataset itself, but instead provide a set of scripts and metadata for other researchers to generate the splits themselves~\footnote{\href{https://github.com/UBC-NLP/arab_voices}{https://github.com/UBC-NLP/arab\_voices}}. The scripts take the downloaded data archives and generate corresponding parquet files with standardized metadata and ISO-aligned language codes.

\section{Dataset Analysis}
\label{sec:analysismethods}

\begin{table*}[]
\resizebox{\textwidth}{!}{%
\begin{tabular}{lrrrrrlrrrlrrrrlr}
\toprule
\multirow{2}{*}{\textbf{Dialect}}                              & \multicolumn{5}{c}{\textbf{Encode-Only (CTC / Acoustic Encoder)}}                                                                                                          &  & \multicolumn{3}{c}{\textbf{Encode-Decoder}}                                                             &  & \multicolumn{6}{c}{\textbf{Multimodal (Speech + LLM Core)}}                                                                                                                                                     \\ \cmidrule{2-6} \cmidrule{8-10} \cmidrule{12-17}

& \multicolumn{1}{c}{\colorbox{blue!15}{\textbf{E1}}} & \multicolumn{1}{c}{\colorbox{blue!15}{\textbf{E2}}} & \multicolumn{1}{c}{\colorbox{blue!15}{\textbf{E3}}} & \multicolumn{1}{c}{\colorbox{blue!15}{\textbf{E4}}} & \multicolumn{1}{c}{\colorbox{blue!15}{\textbf{E5}}} & & \multicolumn{1}{c}{\colorbox{orange!15}{\textbf{ED1}}} & \multicolumn{1}{c}{\colorbox{orange!15}{\textbf{ED2}}} & \multicolumn{1}{c}{\colorbox{orange!15}{\textbf{ED3}}} & & \multicolumn{1}{c}{\colorbox{green!15}{\textbf{M1}}} & \multicolumn{1}{c}{\colorbox{green!15}{\textbf{M2}}} & \multicolumn{1}{c}{\colorbox{green!15}{\textbf{M3}}} & \multicolumn{1}{c}{\colorbox{green!15}{\textbf{M4}}} & \multicolumn{1}{c}{\colorbox{green!15}{\textbf{M5}}} & \multicolumn{1}{c}{\colorbox{green!15}{\textbf{M6}}} \\
\toprule

MSA (arb) & 38.94 & 26.18 & 16.49 & 13.5 & 13.8 &   & 18.51 & 15.55 & 14.7 &   & 17.52 & 11.51 & 13.57 & 11.52 & 11.3 & \textbf{10.19}\\ \midrule

Algerian (arq) & 96.19 & 88.26 & 81.65 & 78.64 & \textbf{78.01} &   & 175.34 & 116.92 & 124.2 &   & 109.76 & 82.56 & 83.6 & 79.25 & 78.92 & 86.29\\
Egyptian (arz) & 74.32 & 58.87 & 45.7 & 42.28 & 45.59 &   & 48.16 & 35.69 & 37.61 &   & 43.25 & 68.6 & 41.7 & 49.38 & 48.76 & \textbf{34.8}\\

Hassaniyya (mey) & 96.57 & 91.79 & 89.19 & 88.55 & 88.45 &   & 204.97 & 135.94 & 93.46 &   & 106.15 & 83.52 & 88.24 & 84.3 & 84.58 & \textbf{83.17}\\

Hijazi (Saudi Arabia) (acw) & 82.03 & 72.77 & 59.08 & 56.73 & 58.05 &   & 191.21 & 105.94 & 62.14 &   & 91.53 & 55.93 & 67.79 & 73.33 & 56.43 & \textbf{53.49}\\

Khaleeji (afb)\textsuperscript{$\star$} & 89.62 & 81.54 & 74.96 & 73.54 & 71.72 &   & 213.89 & 105.14 & 61.89 &   & 80.23 & \textbf{53.71} & 121.57 & 84.05 & 91.34 & 68.39\\

Levantine (apc)\textsuperscript{$\star$} & 51.75 & 38.91 & 29.72 & 26.84 & 26.88 &   & 36.31 & 32.53 & 26.38 &   & 28.28 & 23.81 & 27.61 & 25.45 & 25.17 & \textbf{23.53}\\

Mesopotamian Arabic (Iraq) (acm) & 94.02 & 86.43 & 80.36 & 81.2 & 80.13 &   & 271.25 & 119.91 & 106.26 &   & 87.53 & \textbf{71.16} & 108.36 & 86.76 & 81.93 & 73.86\\ 
Moroccan (ary) & 113.83 & 84.48 & 79.78 & \textbf{79.71} & 81.77 &   & 164.21 & 120.43 & 86.2 &   & 122.79 & 99.33 & 94.68 & 98.85 & 93.19 & 103.8\\
Najdi (Saudi Arabia) (ars) & 84.01 & 76.93 & 64.06 & 59.31 & 60.17 &   & 173.42 & 113.92 & 62.63 &   & 72.73 & \textbf{46.9} & 80.75 & 63.92 & 61.31 & 60.23\\

North Mesopotamian Arabic (Iraq) (ayp) & 94.65 & 89.14 & 87.58 & 87.99 & 84.11 &   & 367.74 & 134.53 & 87.78 &   & 99.13 & \textbf{75.45} & 88.48 & 93.12 & 104.72 & 95.16\\ 

Sanaani Arabic (Yemen) (ayn) & 74.99 & 58.25 & 50.36 & 50.49 & 47.81 &   & 188.3 & 59.06 & 44.25 &   & 47.67 & \textbf{34.63} & 52.32 & 53.7 & 46.19 & 40.37\\
Sudanese (apd) & 74.82 & 63.97 & 49.1 & 45.71 & 44.12 &   & 70.9 & 49.19 & 47.26 &   & 48.71 & 40.06 & 45.49 & 37.68 & 41.33 & \textbf{36.01}\\

Ta'izzi-Adeni Arabic (Yemen) (acq) & 72.61 & 55.22 & 46.42 & 45.84 & 45.44 &   & 125.22 & 77.06 & 39.54 &   & 50.62 & \textbf{28.77} & 44.02 & 43.43 & 38.03 & 35.5\\ 
Tunisian (aeb) & 92.46 & 86.38 & 78.37 & 80.51 & 79.33 &   & 225.04 & 111.93 & 80.63 &   & 85.32 & 247.39 & 80.62 & 84.17 & 77.55 & \textbf{74.31}\\
\bottomrule
\end{tabular}%
}
\caption{WER performance across varieties for 14 models spanning three distinct ASR architectural paradigms on  TEST dataset. \textbf{Encode-Only:} \colorbox{blue!15}{\textbf{E1.}} MMS-1B-ALL, \colorbox{blue!15}{\textbf{E2.}} omniASR-CTC-300M-v2, \colorbox{blue!15}{\textbf{E3.}} omniASR-CTC-1B-v2, \colorbox{blue!15}{\textbf{E4.}} omniASR-CTC-3B-v2, and \colorbox{blue!15}{\textbf{E5.}} omniASR-CTC-7B-v2. \textbf{Encode-Decoder:} \colorbox{orange!15}{\textbf{ED1.}} Whisper-Large-v3-Turbo, \colorbox{orange!15}{\textbf{ED2.}} Whisper-Large-v3, and \colorbox{orange!15}{\textbf{ED3.}} SeamlessM4T-v2-Large. \textbf{Multimodal:} \colorbox{green!15}{\textbf{M1.}} Voxtral-Small-24B-2507, \colorbox{green!15}{\textbf{M2.}} Qwen3-Omni-30B-A3B-Instruct, \colorbox{green!15}{\textbf{M3.}} omniASR-LLM-300M-v2, \colorbox{green!15}{\textbf{M4.}} omniASR-LLM-1B-v2, \colorbox{green!15}{\textbf{M5.}} omniASR-LLM-3B-v2, and \colorbox{green!15}{\textbf{M6.}} omniASR-LLM-7B-v2. \textbf{Bold} refers to the best performance for each dialect. \textsuperscript{$\star$}Khaleeji (afb) and Levantine (apc) include varieties from multiple countries. \textit{For CER performance, details are provided in Table~\ref{appdx_tab:results_cer} and for subdialect breakdown see Table~\ref{tab:appendix_asr_results_wer}.}  
}
\label{tab:asr_results_wer}

\end{table*}

\paragraph{Metadata coverage.}
Across the major dimensions outlined in \S\ref{sec:motivation}, we find that only a subset can be directly recovered from released metadata. We summarize available metadata for dialect coverage, age, gender, and domain in \S\ref{appdx_sec:analysismethods}. Therefore, beyond metadata alone, we rely on automated characterization to support consistent comparison across datasets and to inform benchmark construction.

\paragraph{Text analysis of dialectness.}
Using ALDi, we observe patterns that are consistent with the intended collection conditions of several corpus types. Broadcast news datasets (MASC, QASR, MGB2) are predominantly MSA, and a similar trend holds for many read-speech datasets, including FLEUR, CATTS (Classical Arabic), and TUMSA. For MGB2, the original release estimated a lower bound of at least 70\% MSA \cite{ali2016mgb}. Our ALDi distributions are broadly consistent with this estimate: 24.5\% of utterances have ALDi $>0.11$ (corresponding to \textit{little DA} in \citealp{keleg2023aldi}), while 5.6\% exceed 0.44 (\textit{mixed}) and 1\% exceed 0.77 (\textit{mostly DA}). In aggregate, the \textit{mostly DA} bin corresponds to 11 hours of speech that could potentially be isolated for dialect-focused use.

At the other end of the spectrum, conversational telephone datasets (GULTL, IRQTL, CALLH) are largely dialectal, although YEMTL is a notable exception. Compared to MSA-oriented corpora, these datasets exhibit a wider spread in degree of dialectness. A similar pattern holds for datasets collected explicitly for particular dialects, such as SADA, MGB3, and MGB5. We provide dataset-level violin plots in \S\ref{appdx_sec:analysismethods}. These ALDi-based predictions are mirrored by our binary MSA/DA classifier (Figure~\ref{fig:msa_da}); computing Pearson correlation between the ALDi scores and the binary predictions (MSA$=0$, DA$=1$) yields $r=0.91$. Therefore, we use dialectness distributions as a complementary signal when selecting and interpreting per-dialect adaptation and evaluation subsets in our standardized protocol.

\paragraph{Noise and intelligibility.}

\begin{figure*}[!ht]
    \centering
    \includegraphics[width=0.98\linewidth]{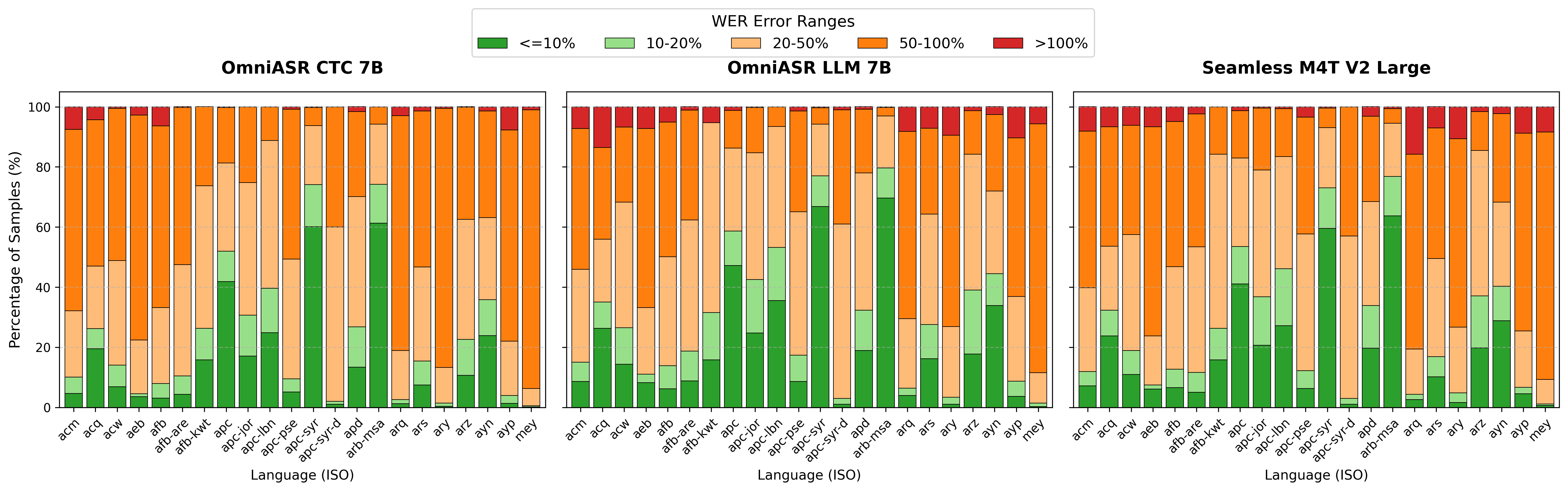}
    \caption{WER distribution across languages for the best-performing ASR models from the three architectures listed in Table~\ref{tab:asr_results_wer}. Stacked bar charts show the proportion of samples in different WER ranges (from $\leq$10\% to $>$100\%) across multiple languages. Colors range from green (low error rates) to red (high error rates). ``\textit{OmniASR LLM 7B}'' consistently achieves a higher share of low-WER samples and more stable performance across languages, highlighting the advantages of LLM-based architectures over CTC-based approaches for multilingual ASR. \textit{For the full analysis on WER and CER performance, details are provided in Figure~\ref{appdx_fig:WER_analysis} and Figure~\ref{appdx_fig:CER_analysis}  Appendix~\ref{appdx_sec:results}.} }
\end{figure*}

We next analyze predicted recording conditions using our audio-quality proxies. Consistent with the recording setups typically used for TTS corpora (Table~\ref{tab:squim}), ADTTS, CATTS, and IRTTS exhibit high predicted quality under the objective proxies (all mean predicted PESQ $>3$ and mean predicted SI-SDR $>20$). Several other read-speech datasets, including ARVOI, TUNAI, and TUMSA, show similarly strong scores. More detailed anaylsis in Appendix \S\ref{appdx_sec:analysismethods}




\section{Experimental Setup}
\label{sec:evaluation}

We evaluate the zero-shot performance of publicly available pretrained models on the development and test splits of our benchmark, reporting both overall results and per-dialect performance. We use Word Error Rate (WER) as the primary metric and Character Error Rate (CER) as a secondary metric for assessing ASR quality. To ensure consistent scoring across heterogeneous corpora, we apply a standardized text normalization pipeline to both model predictions and reference transcripts prior to computing WER/CER; details are provided in \S\ref{appd_subsec:normalization}.

We evaluate the following models: Whisper-large-v3 \cite{radford2023robust}, MMS \cite{pratap2024scaling}, SeamlessM4T v2 \cite{barrault2023seamless}, Omnilingual \cite{omnilingual2025omnilingual}, and Qwen3 \cite{yang2025qwen3}. Model descriptions and configuration details are provided in \S\ref{appdx_sec:evaluation}.
\section{Results and Discussion}
\label{sec:results}

\paragraph{Dataset analysis.}
Our metadata audit highlights a substantial gap between the variability dimensions in our idealized framework (\S\ref{sec:motivation}) and what is routinely recorded in released DA corpora. Utterance-level sociolinguistic descriptors (e.g., register or affect) are uncommon, with emotion recognition datasets being a notable exception where such labels are often integral to the task. We also observe inconsistency in how location metadata is used: in some cases it reflects speaker origin (and thus may index regional accent), while in others it reflects only the recording site, which can confound dialect mapping and downstream interpretation. Demographic fields are similarly uneven: gender is the most frequently recorded attribute, and age is sometimes available, whereas socioeconomic, education, or rural-urban indicators are rarely documented despite their potential value for understanding representativeness. Therefore, our benchmark and standardized protocol rely on harmonized dialect labels where possible and supplement missing metadata with automated dataset characterization.

Automated profiling provides more reliable signals for transcript-based characterization than for recording-condition assessment. While our predicted objective audio-quality proxies are broadly consistent with one another (Figure~\ref{fig:audio_landscape}), they do not consistently track human judgments in our manual inspection. Likewise, the conservative behavior of the non-matching reference MOS predictor may reflect domain mismatch on out-of-domain samples. In qualitative review, we encountered cases where audio that sounded clean received low predicted scores and vice versa, and the relationship between these proxies and downstream model training remains unclear. Therefore, we treat audio-quality estimates as comparative context signals rather than as definitive measures, and we retain dataset provenance in our benchmark/adaptation protocol to support analysis conditioned on recording conditions.

\paragraph{Human verification.}
To contextualize the automated profiling signals, we conduct a human-in-the-loop check comparing model predictions against native-speaker annotations (Appendix~\S\ref{app_subsec:human}). We find that coarse binary distinctions (MSA vs.\ Dialect) and ALDi are both quite accurate in analyzing transcriptions. We further observe that standard quality metrics (e.g., PESQ) can penalize expressive speech (e.g., emotional or religious recordings) that human listeners rate highly, reinforcing the need to interpret predicted quality proxies with caution. Therefore, our analysis layer emphasizes robust, coarse-grained dialectness signals for dataset characterization, and we use audio-quality proxies primarily for contextualizing benchmark outcomes rather than for filtering or ranking data.

\paragraph{Benchmark results.}
Across dialects, zero-shot performance of off-the-shelf models remains challenging (Table~\ref{tab:asr_results_wer}). Nevertheless, we observe several stronger configurations on specific varieties: Omnilingual LLM 7B achieves WERs of 34.8 (arz), 23.53 (apc), and 36.0 (apd), while Qwen3 attains WERs of 34.63 (ayn) and 28.77 (acq). For MSA, we observe that three of four Omnilingual LLM variants and Qwen3 achieve WER below 12.0. Therefore, our standardized evaluation protocol provides a consistent basis for identifying which model families transfer more effectively in a zero-shot setting and where per-dialect adaptation is most warranted.

We further note strong performance from the Omnilingual CTC models on the low-resource Algerian (arq) and Moroccan (ary) subsets (see also CER results in Table~\ref{appdx_tab:results_cer}). We report results without language-model decoding; incorporating an LM could plausibly improve CTC performance, but a practical barrier is obtaining suitably in-domain text for the target variety. Therefore, our benchmark design, with fixed per-dialect adaptation, dev, and test splits, is intended to support controlled exploration of such adaptation strategies (including LM-based approaches) under comparable supervision budgets.

Finally, we observe a characteristic failure mode for encoder-decoder models: in some cases, Whisper produces WER above 200, corresponding to excessively long, low-quality outputs that appear consistent with failures to terminate generation appropriately. Therefore, our benchmark/adaptation protocol, combined with the accompanying analysis layer, facilitates systematic identification of such failure modes across dialects and recording conditions.

\paragraph{Existing gaps.}
A major methodological gap is the lack of an \emph{audio-based} analogue of ALDi \cite{keleg2023aldi}. While text-based dialectness models help characterize corpora with transcripts, their reliance on text prevents direct application to untranscribed speech (e.g., spoken dialect identification settings). In terms of resource coverage, we identify several dialects that remain under-supported and challenging in our benchmark, particularly North African varieties (Algerian, Hassaniyya, Moroccan, Tunisian) and Iraqi varieties (Mesopotamian and North Mesopotamian), and we are unable to identify datasets for several other dialects within our cutoff and inclusion criteria (Table~\ref{tab:arabic_dialects_main}). We also reiterate the need for dataset creators to document collection conditions more systematically and to expand coverage toward underrepresented domains (e.g., health, arts, and nature). Therefore, \benchmark provides both a reproducible evaluation protocol and a practical scaffold for prioritizing future data collection and dialect adaptation efforts where gaps are most acute.
\section{Conclusion}
\label{sec:conclusion}
We present a detailed analysis of a broad selection of spoken Arabic datasets through the complementary lenses of linguistic \emph{dialectness} and recording conditions. We standardize and map heterogeneous resources into a unified schema, and we introduce a benchmark with a standardized evaluation and training/adaptation protocol for assessing modern ASR performance on Dialectal Arabic. Our analysis highlights substantial variation across corpora and underscores persistent gaps in both documentation and coverage. We identify several directions for future work: (1) evaluating how predicted audio-quality measures relate to ASR performance in practice; (2) expanding benchmark data to cover additional low-resource dialects; (3) developing language models suitable for CTC decoding in Dialectal Arabic; and (4) extending benchmarks to a broader range of domains.

\section*{Limitations}
\label{sec:limitations}
Our dialect mapping uses paired ISO~639-3 labels and country-region tags, which can support more fine-grained organization than country-only labeling. However, this representation alone may not capture distinctions that cut across geography, such as rural-urban variation, and further refinement would require additional metadata that is often unavailable in released corpora.

In addition, ISO labels are an imperfect administrative proxy for a dialect continuum: they may not fully reflect sociolinguistic realities and can be revised over time, which may complicate long-term reproducibility of mappings. Our analysis and benchmark should therefore be interpreted as a best-effort snapshot under the dataset releases and labeling conventions available within our collection window.

Finally, our dialectness analysis relies on text-based models such as ALDi, which can miss pronunciation differences that are not reflected orthographically (e.g., cases where written DA and MSA are indistinguishable without diacritization). Developing an audio-based analogue of ALDi (``spoken ALDi'') is a promising direction to address this limitation in future work.

\section*{Ethical Considerations}
Our framework emphasizes sociolinguistic and contextual factors that can matter for real-world speech technology, but richer documentation can also increase privacy risks for recording participants. In particular, inferring sensitive attributes from audio or transcripts (e.g., gender, socioeconomic status, or education) can be intrusive and may enable unwanted profiling. Accordingly, our work focuses on dialect and recording-condition characterization and reports results in aggregate, and we do not attempt to infer sensitive demographic attributes beyond what is explicitly provided in released metadata. We encourage dataset creators to balance transparency with participant privacy and to document consent, intended use, and data handling practices when possible.

\section*{Acknowledgments}\label{sec:acknow}
We acknowledge support from Canada Research Chairs (CRC), the Natural Sciences and Engineering Research Council of Canada (NSERC; RGPIN-2018-04267), the Social Sciences and Humanities Research Council of Canada (SSHRC; 895-2020-1004), Canadian Foundation for Innovation (CFI; 37771), Digital Research Alliance of Canada,\footnote{\href{https://alliancecan.ca}{https://alliancecan.ca}} and UBC ARC-Sockeye.\footnote{\href{https://arc.ubc.ca/ubc-arc-sockeye}{https://arc.ubc.ca/ubc-arc-sockeye}} 

\normalem
\bibliography{custom}

@inproceedings{al2023masc,
  title={MASC: Massive Arabic speech corpus},
  author={Al-Fetyani, Mohammad and Al-Barham, Muhammad and Abandah, Gheith and Alsharkawi, Adham and Dawas, Maha},
  booktitle={2022 IEEE Spoken Language Technology Workshop (SLT)},
  pages={1006--1013},
  year={2023},
  organization={IEEE}
}

@inproceedings{ali2017speech,
  title={Speech recognition challenge in the wild: Arabic MGB-3},
  author={Ali, Ahmed and Vogel, Stephan and Renals, Steve},
  booktitle={2017 IEEE Automatic Speech Recognition and Understanding Workshop (ASRU)},
  pages={316--322},
  year={2017},
  organization={IEEE}
}

@inproceedings{ali2019mgb,
  title={The mgb-5 challenge: Recognition and dialect identification of dialectal arabic speech},
  author={Ali, Ahmed and Shon, Suwon and Samih, Younes and Mubarak, Hamdy and Abdelali, Ahmed and Glass, James and Renals, Steve and Choukri, Khalid},
  booktitle={2019 IEEE Automatic Speech Recognition and Understanding Workshop (ASRU)},
  pages={1026--1033},
  year={2019},
  organization={IEEE}
}

@phdthesis{halabi2016modern,
  title={Modern standard Arabic phonetics for speech synthesis},
  author={Halabi, Nawar},
  year={2016},
  school={University of Southampton}
}

@inproceedings{kulkarni2023clartts,
  author={Ajinkya Kulkarni and Atharva Kulkarni and Sara Abedalmon'em Mohammad Shatnawi and Hanan Aldarmaki},
  title={ClArTTS: An Open-Source Classical Arabic Text-to-Speech Corpus},
  year={2023},
  booktitle={2023 INTERSPEECH },
  pages={5511--5515},
  doi={10.21437/Interspeech.2023-2224}
}

@misc{toyin2025arvoicemultispeakerdatasetarabic,
      title={ArVoice: A Multi-Speaker Dataset for Arabic Speech Synthesis}, 
      author={Hawau Olamide Toyin and Rufael Marew and Humaid Alblooshi and Samar M. Magdy and Hanan Aldarmaki},
      year={2025},
      eprint={2505.20506},
      archivePrefix={arXiv},
      primaryClass={cs.CL},
      url={https://arxiv.org/abs/2505.20506}, 
}

@misc{kharrufa_2024_11170567,
  author       = {Kharrufa, Hayder and
                  Taha, Adam and
                  Baraq, Mohammed},
  title        = {{Training a Text-to-Speech System for Dialectal 
                   Arabic with a Focus on the Iraqi Dialect}},
  month        = may,
  year         = 2024,
  publisher    = {Zenodo},
  version      = {1.0},
  doi          = {10.5281/zenodo.11170567},
  url          = {https://doi.org/10.5281/zenodo.11170567}
}

@inproceedings{hamed2020arzen,
  title={ArzEn: A speech corpus for code-switched Egyptian Arabic-English},
  author={Hamed, Injy and Vu, Ngoc Thang and Abdennadher, Slim},
  booktitle={Proceedings of the twelfth language resources and evaluation conference},
  pages={4237--4246},
  year={2020}
}

@inproceedings{mixat,
  title={Mixat: A Data Set of Bilingual Emirati-English Speech},
  author={Al Ali, Maryam and Aldarmaki, Hanan},
  booktitle={SIGUL 2024: 3rd Annual Meeting of the ELRA/ISCA Special Interest Group on Under-resourced Languages, a Satellite Workshop of LREC-COLING 2024},
  year={2024}
}

@misc{SCC2025,
  title={Saudilang Code-Switch Corpus (SCC)},
  author={SDAIA},
  year={2022},
  howpublished={\url{https://www.kaggle.com/datasets/sdaiancai/saudilang-code-switch-corpus-scc}},
  note={CC BY-NC-SA 4.0}
}

@inproceedings{habash2022zaebuc,
  title={ZAEBUC: An annotated Arabic-English bilingual writer corpus},
  author={Habash, Nizar and Palfreyman, David},
  booktitle={Proceedings of the Thirteenth Language Resources and Evaluation Conference},
  pages={79--88},
  year={2022}
}

@misc{LDC2023S10_kasdi,
title={
Kasdi-Merbah University Emotional Database in Arabic Speech},
author={Mourad Belhadj and Ilham Bendellali and Elalia Lakhdari},
year={2023},
howpublished={\url{https://doi.org/10.35111/qqer-qz15}},
note={LDC2023S10}

}

@inproceedings{conneau2023fleurs,
  title={Fleurs: Few-shot learning evaluation of universal representations of speech},
  author={Conneau, Alexis and Ma, Min and Khanuja, Simran and Zhang, Yu and Axelrod, Vera and Dalmia, Siddharth and Riesa, Jason and Rivera, Clara and Bapna, Ankur},
  booktitle={2022 IEEE Spoken Language Technology Workshop (SLT)},
  pages={798--805},
  year={2023},
  organization={IEEE}
}

@inproceedings{ali2016mgb,
  title={The MGB-2 challenge: Arabic multi-dialect broadcast media recognition},
  author={Ali, Ahmed and Bell, Peter and Glass, James and Messaoui, Yacine and Mubarak, Hamdy and Renals, Steve and Zhang, Yifan},
  booktitle={2016 IEEE Spoken Language Technology Workshop (SLT)},
  pages={279--284},
  year={2016},
  organization={IEEE}
}

@article{talafha2024casablanca,
  title={Casablanca: Data and Models for Multidialectal Arabic Speech Recognition},
  author={Talafha, Bashar and Kadaoui, Karima and Magdy, Samar Mohamed and Habiboullah, Mariem and Chafei, Chafei Mohamed and El-Shangiti, Ahmed Oumar and Zayed, Hiba and Alhamouri, Rahaf and Assi, Rwaa and Alraeesi, Aisha and others},
  journal={arXiv preprint arXiv:2410.04527},
  year={2024}
}

@inproceedings{alharbi2024sada,
  title={Sada: Saudi audio dataset for arabic},
  author={Alharbi, Sadeen and Alowisheq, Areeb and T{\"u}ske, Zolt{\'a}n and Darwish, Kareem and Alrajeh, Abdullah and Alrowithi, Abdulmajeed and Tamran, Aljawharah Bin and Ibrahim, Asma and Aloraini, Raghad and Alnajim, Raneem and others},
  booktitle={ICASSP 2024-2024 IEEE International Conference on Acoustics, Speech and Signal Processing (ICASSP)},
  pages={10286--10290},
  year={2024},
  organization={IEEE}
}

@article{mubarak2021qasr,
  title={QASR: QCRI Aljazeera Speech Resource--A Large Scale Annotated Arabic Speech Corpus},
  author={Mubarak, Hamdy and Hussein, Amir and Chowdhury, Shammur Absar and Ali, Ahmed},
  journal={arXiv preprint arXiv:2106.13000},
  year={2021}
}

@article{ardila2019common,
  title={Common voice: A massively-multilingual speech corpus},
  author={Ardila, Rosana and Branson, Megan and Davis, Kelly and Henretty, Michael and Kohler, Michael and Meyer, Josh and Morais, Reuben and Saunders, Lindsay and Tyers, Francis M and Weber, Gregor},
  journal={arXiv preprint arXiv:1912.06670},
  year={2019}
}

@inproceedings{garnerinInvestigatingImpactGender2021,
  title = {Investigating the {{Impact}} of {{Gender Representation}} in {{ASR Training Data}}: A {{Case Study}} on {{Librispeech}}},
  shorttitle = {Investigating the {{Impact}} of {{Gender Representation}} in {{ASR Training Data}}},
  booktitle = {Proceedings of the 3rd {{Workshop}} on {{Gender Bias}} in {{Natural Language Processing}}},
  author = {Garnerin, Mahault and Rossato, Solange and Besacier, Laurent},
  year = {2021},
  pages = {86--92},
  publisher = {Association for Computational Linguistics},
  location = {Online},
  doi = {10.18653/v1/2021.gebnlp-1.10},
  url = {https://aclanthology.org/2021.gebnlp-1.10},
  urldate = {2024-06-28},
  eventtitle = {Proceedings of the 3rd {{Workshop}} on {{Gender Bias}} in {{Natural Language Processing}}},
  langid = {english},
}

@inproceedings{soltau2011modern,
  title={From modern standard arabic to levantine asr: Leveraging gale for dialects},
  author={Soltau, Hagen and Mangu, Lidia and Biadsy, Fadi},
  booktitle={2011 IEEE Workshop on Automatic Speech Recognition \& Understanding},
  pages={266--271},
  year={2011},
  organization={IEEE}
}

@article{alsharhan2020investigating,
  title={Investigating the effects of gender, dialect, and training size on the performance of Arabic speech recognition},
  author={Alsharhan, Eiman and Ramsay, Allan},
  journal={Language Resources and Evaluation},
  volume={54},
  number={4},
  pages={975--998},
  year={2020},
  publisher={Springer}
}

@misc{LDC2013S07_GALE_p2_conv_1,
title={GALE Phase 2 Arabic Broadcast Conversation Speech Part 1
},
  year={2013},

author={ 	Kevin Walker and Christopher Caruso and Kazuaki Maeda and Denise DiPersio and Stephanie Strassel},
howpublished={\url{https://doi.org/10.35111/j224-tx54}},
note={LDC2013S07}
}

@misc{LDC97S45_callhome_egypt,
title={CALLHOME Egyptian Arabic Speech
},
  year={1997},
author={Alexandra Canavan and George Zipperlen and David Graff},
howpublished={\url{https://doi.org/10.35111/d8yb-9m13}},
note={LDC97S45}
}

@misc{LDC2024S11_l2_ksu,
title={L2-KSU Native and Non-Native Arabic Speech
},
  year={2024},
author={Norah Alrashoudi and Hend AlKhalifa and Yousef Ajami Alotaibi},
howpublished={\url{https://doi.org/10.35111/n3d8-t960}},
note={LDC2024S11}
}

@misc{LDC2006S45_iraqi_speech,
title={Iraqi Arabic Conversational Telephone Speech},
author={{Appen Pty Ltd}},
  year={2006},

howpublished={\url{https://doi.org/10.35111/2dcs-9751}},
note={LDC2006S45}
}

@misc{LDC2006S43_gulf_speech,
title={Gulf Arabic Conversational Telephone Speech},
author={{Appen Pty Ltd}},
  year={2006},
howpublished={\url{https://doi.org/10.35111/nsvg-dd69}},
note={LDC2006S43}
}

@misc{LDC2017S12_KSUEmotions,
title={KSUEmotions},
author={Ali Hamid Meftah and Yousef Ajami Alotaibi and Sid-Ahmed Selouani},
  year={2017},
howpublished={\url{https://doi.org/10.35111/q1eh-6457}},
note={LDC2017S12}
}

@misc{magicdata_no_date_egyconv,
title={ASR-EgArbCSC: An Egyptian Arabic Conversational Speech Corpus},
author={MagicData},
  year={no date},
url="https://magichub.com/datasets/egyptian-arabic-conversational-speech-corpus/",
}

@misc{magicdata_no_date_yemconv,
title={ASR-EgArbCSC: An Egyptian Arabic Conversational Speech Corpus},
author={MagicData},
  year={no date},
url="https://magichub.com/datasets/yemeni-arabic-conversational-speech-corpus/",
}

@article{alsayadi2022deep,
  title={Deep investigation of the recent advances in dialectal arabic speech recognition},
  author={Alsayadi, Hamzah A and Abdelhamid, Abdelaziz A and Hegazy, Islam and Alotaibi, Bandar and Fayed, Zaki T},
  journal={IEEE access},
  volume={10},
  pages={57063--57079},
  year={2022},
  publisher={IEEE}
}

@article{keleg2023aldi,
  title={ALDi: Quantifying the Arabic level of dialectness of text},
  author={Keleg, Amr and Goldwater, Sharon and Magdy, Walid},
  journal={arXiv preprint arXiv:2310.13747},
  year={2023}
}

@inproceedings{abdul2021arbert,
  title={ARBERT \& MARBERT: Deep Bidirectional Transformers for Arabic},
  author={Abdul-Mageed, Muhammad and Elmadany, Abdelrahim and others},
  booktitle={Proceedings of the 59th Annual Meeting of the Association for Computational Linguistics and the 11th International Joint Conference on Natural Language Processing (Volume 1: Long Papers)},
  pages={7088--7105},
  year={2021}
}

@article{alqadasi2023modern,
  title={Modern standard Arabic speech corpora: A systematic review},
  author={Alqadasi, Ammar Mohammed Ali and Abdulghafor, Rawad and Sunar, Mohd Shahrizal and Salam, Md Sah Bin HJ},
  journal={Ieee Access},
  volume={11},
  pages={55771--55796},
  year={2023},
  publisher={IEEE}
}

@inproceedings{radford2023robust,
  title={Robust speech recognition via large-scale weak supervision},
  author={Radford, Alec and Kim, Jong Wook and Xu, Tao and Brockman, Greg and McLeavey, Christine and Sutskever, Ilya},
  booktitle={International conference on machine learning},
  pages={28492--28518},
  year={2023},
  organization={PMLR}
}

@article{goyal2022flores,
	title={The flores-101 evaluation benchmark for low-resource and multilingual machine translation},
	author={Goyal, Naman and Gao, Cynthia and Chaudhary, Vishrav and Chen, Peng-Jen and Wenzek, Guillaume and Ju, Da and Krishnan, Sanjana and Ranzato, Marc’Aurelio and Guzm{\'a}n, Francisco and Fan, Angela},
	journal={Transactions of the Association for Computational Linguistics},
	volume={10},
	pages={522--538},
	year={2022},
	publisher={MIT Press One Broadway, 12th Floor, Cambridge, Massachusetts 02142, USA~…}
}

@misc{slr46,
title={{Tunisian\_MSA}},
author={OpenSLR},
howpublished={https://www.openslr.org/46/},
year={2003}
}

@misc{mediaspeech2021,
      title={MediaSpeech: Multilanguage ASR Benchmark and Dataset}, 
      author={Rostislav Kolobov and Olga Okhapkina and Olga Omelchishina, Andrey Platunov and Roman Bedyakin and Vyacheslav Moshkin and Dmitry Menshikov and Nikolay Mikhaylovskiy},
      year={2021},
      eprint={2103.16193},
      archivePrefix={arXiv},
      primaryClass={eess.AS}
}

@misc{slr132,
	title={{Mohammed - Quran Speech to Text Dataset}},
	author={OpenSLR},
	howpublished={https://www.openslr.org/132/},
	year={}
}

@article{albirini2011sociolinguistic,
  title={The sociolinguistic functions of codeswitching between Standard Arabic and Dialectal Arabic},
  author={Albirini, Abdulkafi},
  journal={Language in society},
  volume={40},
  number={5},
  pages={537--562},
  year={2011},
  publisher={Cambridge University Press}
}

@inproceedings{hamed-etal-2024-zaebuc,
    title = "{ZAEBUC}-Spoken: A Multilingual Multidialectal {A}rabic-{E}nglish Speech Corpus",
    author = "Hamed, Injy  and
      Eryani, Fadhl  and
      Palfreyman, David  and
      Habash, Nizar",
    editor = "Calzolari, Nicoletta  and
      Kan, Min-Yen  and
      Hoste, Veronique  and
      Lenci, Alessandro  and
      Sakti, Sakriani  and
      Xue, Nianwen",
    booktitle = "Proceedings of the 2024 Joint International Conference on Computational Linguistics, Language Resources and Evaluation (LREC-COLING 2024)",
    month = may,
    year = "2024",
    address = "Torino, Italia",
    publisher = "ELRA and ICCL",
    url = "https://aclanthology.org/2024.lrec-main.1546/",
    pages = "17770--17782"
}

@article{alqadasi2025arabic,
  title={Arabic Dialects Speech Corpora: A Systematic Review},
  author={Alqadasi, Ammar Mohammed Ali and Zeki, Akram M and Sunar, Mohd Shahrizal and Hashim, Siti Zaiton Mohd and hj Salam, Md Sah and Abdulghafor, Rawad},
  journal={Speech Communication},
  pages={103322},
  year={2025},
  publisher={Elsevier}
}

@article{bender2018data,
  title={Data statements for natural language processing: Toward mitigating system bias and enabling better science},
  author={Bender, Emily M and Friedman, Batya},
  journal={Transactions of the Association for Computational Linguistics},
  volume={6},
  pages={587--604},
  year={2018},
  publisher={MIT Press One Rogers Street, Cambridge, MA 02142-1209, USA journals-info~…}
}

@inproceedings{agostinelli-etal-2025-findings,
    title = "Findings of the {IWSLT} 2025 Evaluation Campaign",
    author = {Abdulmumin, Idris  and
      Agostinelli, Victor  and
      Alum{\"a}e, Tanel  and
      Anastasopoulos, Antonios  and
      Bentivogli, Luisa  and
      Bojar, Ond{\v{r}}ej  and
      Borg, Claudia  and
      Bougares, Fethi  and
      Cattoni, Roldano  and
      Cettolo, Mauro  and
      Chen, Lizhong  and
      Chen, William  and
      Dabre, Raj  and
      Est{\`e}ve, Yannick  and
      Federico, Marcello  and
      Fishel, Mark  and
      Gaido, Marco  and
      Javorsk{\'y}, D{\'a}vid  and
      Kasztelnik, Marek  and
      Kponou, Fortun{\'e}  and
      Krubi{\'n}ski, Mateusz  and
      Kin Lam, Tsz  and
      Liu, Danni  and
      Matusov, Evgeny  and
      Kumar Maurya, Chandresh  and
      P. McCrae, John  and
      Mdhaffar, Salima  and
      Moslem, Yasmin  and
      Murray, Kenton  and
      Nakamura, Satoshi  and
      Negri, Matteo  and
      Niehues, Jan  and
      Kr. Ojha, Atul  and
      Ortega, John E.  and
      Papi, Sara  and
      Pecina, Pavel  and
      Pol{\'a}k, Peter  and
      Po{\l}e{\'c}, Piotr  and
      Sankar, Ashwin  and
      Savoldi, Beatrice  and
      Sethiya, Nivedita  and
      Sikasote, Claytone  and
      Sperber, Matthias  and
      St{\"u}ker, Sebastian  and
      Sudoh, Katsuhito  and
      Thompson, Brian  and
      Turchi, Marco  and
      Waibel, Alex  and
      Wilken, Patrick  and
      Zevallos, Rodolfo  and
      Zouhar, Vil{\'e}m  and
      Z{\"u}fle, Maike},
    editor = "Salesky, Elizabeth  and
      Federico, Marcello  and
      Anastasopoulos, Antonis",
    booktitle = "Proceedings of the 22nd International Conference on Spoken Language Translation (IWSLT 2025)",
    month = jul,
    year = "2025",
    address = "Vienna, Austria (in-person and online)",
    publisher = "Association for Computational Linguistics",
    url = "https://aclanthology.org/2025.iwslt-1.44/",
    doi = "10.18653/v1/2025.iwslt-1.44",
    pages = "412--481",
    ISBN = "979-8-89176-272-5"
}

@article{omnilingual2025omnilingual,
  title={Omnilingual ASR: Open-Source Multilingual Speech Recognition for 1600+ Languages},
  author={Omnilingual, ASR and Keren, Gil and Kozhevnikov, Artyom and Meng, Yen and Ropers, Christophe and Setzler, Matthew and Wang, Skyler and Adebara, Ife and Auli, Michael and Balioglu, Can and others},
  journal={arXiv preprint arXiv:2511.09690},
  year={2025}
}

@article{pratap2024scaling,
  title={Scaling speech technology to 1,000+ languages},
  author={Pratap, Vineel and Tjandra, Andros and Shi, Bowen and Tomasello, Paden and Babu, Arun and Kundu, Sayani and Elkahky, Ali and Ni, Zhaoheng and Vyas, Apoorv and Fazel-Zarandi, Maryam and others},
  journal={Journal of Machine Learning Research},
  volume={25},
  number={97},
  pages={1--52},
  year={2024}
}

@article{barrault2023seamless,
  title={Seamless: Multilingual Expressive and Streaming Speech Translation},
  author={Barrault, Lo{\"\i}c and Chung, Yu-An and Meglioli, Mariano Coria and Dale, David and Dong, Ning and Duppenthaler, Mark and Duquenne, Paul-Ambroise and Ellis, Brian and Elsahar, Hady and Haaheim, Justin and others},
  journal={arXiv preprint arXiv:2312.05187},
  year={2023}
}

@inproceedings{kumar2023torchaudio,
  title={Torchaudio-squim: Reference-less speech quality and intelligibility measures in torchaudio},
  author={Kumar, Anurag and Tan, Ke and Ni, Zhaoheng and Manocha, Pranay and Zhang, Xiaohui and Henderson, Ethan and Xu, Buye},
  booktitle={ICASSP 2023-2023 IEEE International Conference on Acoustics, Speech and Signal Processing (ICASSP)},
  pages={1--5},
  year={2023},
  organization={IEEE}
}

@inproceedings{rix2001perceptual,
  title={Perceptual evaluation of speech quality (PESQ)-a new method for speech quality assessment of telephone networks and codecs},
  author={Rix, Antony W and Beerends, John G and Hollier, Michael P and Hekstra, Andries P},
  booktitle={2001 IEEE international conference on acoustics, speech, and signal processing. Proceedings (Cat. No. 01CH37221)},
  volume={2},
  pages={749--752},
  year={2001},
  organization={IEEE}
}

@article{taal2011algorithm,
  title={An algorithm for intelligibility prediction of time--frequency weighted noisy speech},
  author={Taal, Cees H and Hendriks, Richard C and Heusdens, Richard and Jensen, Jesper},
  journal={IEEE Transactions on audio, speech, and language processing},
  volume={19},
  number={7},
  pages={2125--2136},
  year={2011},
  publisher={IEEE}
}

@inproceedings{le2019sdr,
  title={SDR--half-baked or well done?},
  author={Le Roux, Jonathan and Wisdom, Scott and Erdogan, Hakan and Hershey, John R},
  booktitle={ICASSP 2019-2019 IEEE International Conference on Acoustics, Speech and Signal Processing (ICASSP)},
  pages={626--630},
  year={2019},
  organization={IEEE}
}

@article{rosli2018evaluating,
  title={Evaluating the quality of datasets in software engineering},
  author={Rosli, Marshima Mohd and Tempero, Ewan and Luxton-Reilly, Andrew},
  journal={Advanced Science Letters},
  volume={24},
  number={10},
  pages={7232--7239},
  year={2018},
  publisher={American Scientific Publishers}
}

@inproceedings{manocha22c_interspeech,
  title     = {{Speech Quality Assessment through MOS using Non-Matching References}},
  author    = {{Pranay Manocha and Anurag Kumar}},
  year      = {{2022}},
  booktitle = {{Interspeech 2022}},
  pages     = {{654--658}},
  doi       = {{10.21437/Interspeech.2022-407}},
  issn      = {{2958-1796}},
}

@inproceedings{naouaralinto,
  title={LinTO Audio and Textual Datasets to Train and Evaluate Automatic Speech Recognition in Tunisian Arabic Dialect},
year={2025},
  author={Naouara, Hedi and Lorr{\'e}, Jean-Pierre and Louradour, J{\'e}r{\^o}me},
  booktitle={Workshop on Preparing Good Data for Generative AI: Challenges and Approaches}
}

@inproceedings{mdhaffar2024taric,
  title={TARIC-SLU: A Tunisian benchmark dataset for spoken language understanding},
  author={Mdhaffar, Salima and Bougares, Fethi and De Mori, Renato and Zaiem, Salah and Ravanelli, Mirco and Est{\`e}ve, Yannick},
  booktitle={Proceedings of the 2024 Joint International Conference on Computational Linguistics, Language Resources and Evaluation (LREC-COLING 2024)},
  pages={15606--15616},
  year={2024}
}

@article{elnagar2021systematic,
  title={Systematic literature review of dialectal Arabic: identification and detection},
  author={Elnagar, Ashraf and Yagi, Sane M and Nassif, Ali Bou and Shahin, Ismail and Salloum, Said A},
  journal={IEEE Access},
  volume={9},
  pages={31010--31042},
  year={2021},
  publisher={IEEE}
}

@article{chemnad2023advancements,
  title={Advancements in Arabic Text-to-speech systems: a 22-year literature review},
  author={Chemnad, Khansa and Othman, Achraf},
  journal={IEEE Access},
  volume={11},
  pages={30929--30954},
  year={2023},
  publisher={IEEE}
}

@inproceedings{sullivan23_interspeech,
  title     = {On the Robustness of Arabic Speech Dialect Identification},
  author    = {Peter Sullivan and AbdelRahim Elmadany and Muhammad Abdul-Mageed},
  year      = {2023},
  booktitle = {Interspeech 2023},
  pages     = {5326--5330},
  doi       = {10.21437/Interspeech.2023-1005},
  issn      = {2958-1796},
}

@inproceedings{abdullah25_interspeech,
  title     = {{Voice Conversion Improves Cross-Domain Robustness  for Spoken Arabic Dialect Identification}},
  author    = {Badr M. Abdullah and Matthew Baas and Bernd Möbius and Dietrich Klakow},
  year      = {2025},
  booktitle = {{Interspeech 2025}},
  pages     = {2790--2794},
  doi       = {10.21437/Interspeech.2025-1809},
  issn      = {2958-1796},
}

@inproceedings{manocha22_interspeech,
  title     = {{Audio Similarity is Unreliable as a Proxy for Audio Quality}},
  author    = {{Pranay Manocha and Zeyu Jin and Adam Finkelstein}},
  year      = {{2022}},
  booktitle = {{Interspeech 2022}},
  pages     = {{3553--3557}},
  doi       = {{10.21437/Interspeech.2022-405}},
  issn      = {{2958-1796}},
}

@article{mysore2014can,
  title={Can we automatically transform speech recorded on common consumer devices in real-world environments into professional production quality speech?—a dataset, insights, and challenges},
  author={Mysore, Gautham J},
  journal={IEEE Signal Processing Letters},
  volume={22},
  number={8},
  pages={1006--1010},
  year={2014},
  publisher={IEEE}
}

@article{likhomanenko2020rethinking,
  title={Rethinking evaluation in ASR: Are our models robust enough?},
  author={Likhomanenko, Tatiana and Xu, Qiantong and Pratap, Vineel and Tomasello, Paden and Kahn, Jacob and Avidov, Gilad and Collobert, Ronan and Synnaeve, Gabriel},
  journal={arXiv preprint arXiv:2010.11745},
  year={2020}
}

@misc{Eberhard2025ethnologue,
  title={Ethnologue: Languages of the world},
  author={Eberhard, David M. and Simons, Gary F. and  Fennig, Charles D.},
  year={2025},
notes={\url{https://www.ethnologue.com/}}
}

@book{wardhaugh2021introduction,
  title={An introduction to sociolinguistics},
  author={Wardhaugh, Ronald and Fuller, Janet M},
  year={2021},
  publisher={John Wiley \& Sons}
}

@inbook{Eckert_2016, place={Cambridge}, title={Variation, meaning and social change}, booktitle={Sociolinguistics: Theoretical Debates}, publisher={Cambridge University Press}, author={Eckert, Penelope}, year={2016}, pages={68–85}}

@inbook{lee2025language,
title={Sociolinguistic variation },
  booktitle={Language in Society},
  author={Lee, Nala H},
  year={2025},
  publisher={Routledge},
chapter={2}
}

@inbook{irvine2002style,
title={“Style” as distinctiveness: the culture and ideology of linguistic differentiation},
chapter={1},
  booktitle={Style and sociolinguistic variation},
  author={Judith T. Irvine},
  year={2001},
  publisher={Cambridge University Press}
}

@article{vaswani2017attention,
  title={Attention is all you need},
  author={Vaswani, Ashish and Shazeer, Noam and Parmar, Niki and Uszkoreit, Jakob and Jones, Llion and Gomez, Aidan N and Kaiser, {\L}ukasz and Polosukhin, Illia},
  journal={Advances in neural information processing systems},
  volume={30},
  year={2017}
}

@inproceedings{gulati20_interspeech,
  title     = {Conformer: Convolution-augmented Transformer for Speech Recognition},
  author    = {Anmol Gulati and James Qin and Chung-Cheng Chiu and Niki Parmar and Yu Zhang and Jiahui Yu and Wei Han and Shibo Wang and Zhengdong Zhang and Yonghui Wu and Ruoming Pang},
  year      = {2020},
  booktitle = {Interspeech 2020},
  pages     = {5036--5040},
  doi       = {10.21437/Interspeech.2020-3015},
  issn      = {2958-1796},
}

@inproceedings{khokhlov24_interspeech,
  title     = {{Classification of Room Impulse Responses and its application for channel verification and diarization}},
  author    = {Yuri Khokhlov and Tatiana Prisyach and Anton Mitrofanov and Dmitry Dutov and Igor Agafonov and Tatiana Timofeeva and Aleksei Romanenko and Maxim Korenevsky},
  year      = {2024},
  booktitle = {{Interspeech 2024}},
  pages     = {3250--3254},
  doi       = {10.21437/Interspeech.2024-2081},
  issn      = {2958-1796},
}

@inproceedings{ryu25b_interspeech,
  title     = {{Unified Microphone Conversion: Many-to-Many Device Mapping via Feature-wise Linear Modulation}},
  author    = {Myeonghoon Ryu and Hongseok Oh and Suji Lee and Han Park},
  year      = {2025},
  booktitle = {{Interspeech 2025}},
  pages     = {1333--1337},
  doi       = {10.21437/Interspeech.2025-1356},
  issn      = {2958-1796},
}

@inproceedings{ferrofilho25_interspeech,
  title     = {{Evaluating Deep Speaker Embedding Robustness to Domain, Sampling Rate, and Codec Variations}},
  author    = {Alexandre {Ferro Filho} and Diogo {Fernandes Costa Silva} and Pedro Elias {Engelberg Silva Borges} and Arlindo Rodrigues {Galvão Filho}},
  year      = {2025},
  booktitle = {{Interspeech 2025}},
  pages     = {1113--1117},
  doi       = {10.21437/Interspeech.2025-2167},
  issn      = {2958-1796},
}

@inproceedings{zaidan2011arabic,
  title={The arabic online commentary dataset: an annotated dataset of informal arabic with high dialectal content},
  author={Zaidan, Omar and Callison-Burch, Chris},
  booktitle={Proceedings of the 49th Annual Meeting of the Association for Computational Linguistics: Human Language Technologies},
  pages={37--41},
  year={2011}
}

@article{yang2025qwen3,
  title={Qwen3 technical report},
  author={Yang, An and Li, Anfeng and Yang, Baosong and Zhang, Beichen and Hui, Binyuan and Zheng, Bo and Yu, Bowen and Gao, Chang and Huang, Chengen and Lv, Chenxu and others},
  journal={arXiv preprint arXiv:2505.09388},
  year={2025}
}

@misc{NourhannADTTS,
year={2025},
author={Nourhan Mahmoud},
title= {Arabic-Diacritized-TTS},
url= {https://huggingface.co/datasets/Nourhann/Arabic-Diacritized-TTS
}}

\clearpage
\appendix
\appendixpage            
\addappheadtotoc         
\numberwithin{figure}{section}
\numberwithin{table}{section}
\numberwithin{subsection}{section}

This appendix provides supplementary material to support the main findings of this work. It is organized as follows:

\begin{itemize}
    \item Appendix \ref{appdx_sec:framework} Framework of Audio Variability
    \item Appendix \ref{appdx_sec:lit_review} Literature Review
    \item Appendix \ref{appdx_sec:mapping} Mapping and Standardization
    \item Appendix \ref{appdx_sec:analysismethods} Dataset Analysis
    \item Appendix \ref{appdx_sec:evaluation} Experimental Setup
    \item Appendix \ref{appdx_sec:results} Results

\end{itemize}

\paragraph{Key Tables}
\begin{itemize}

\item Table~\ref{tab:ideal_framework}: A framework for characterizing aspects of variabiltiy in audio recordings.

\item Table~\ref{tab:arabic_dialects_main}: Dialect-wise breakdown of transcribed Arabic audio by train/dev/test splits (hours and utterances). 
\item Table~\ref{tab:squim}: Audio metrics overview. 
\item Table~\ref{tab:benchmark}: Benchmark overview.

\item Table~\ref{tab:appendix_asr_results_wer}: A comprehensive comparison of WER performance  on TEST dataset.

    \item Table~\ref{appdx_tab:results_cer}: A comprehensive comparison of CER performance  on TEST dataset.

    \item Table~\ref{appdx_tab:results_wer_dev}: A comprehensive comparison of WER performance  on DEV dataset.

\end{itemize}

\paragraph{Key Figures}

\begin{figure*}[h!]
    \centering
    \includegraphics[width=0.9\textwidth]{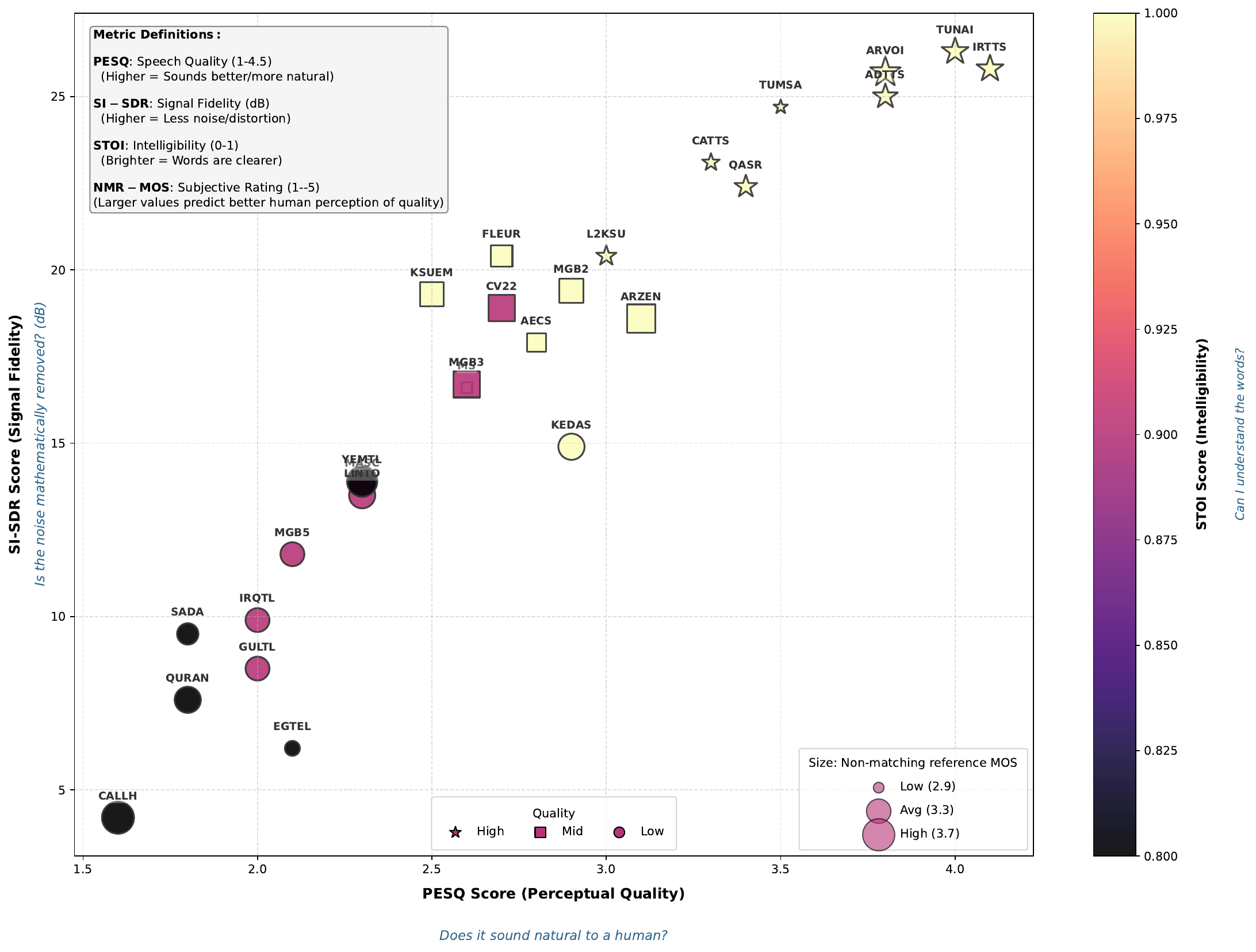}
    
    \caption[Audio Performance Landscape]{We plot the relationship between the four audio quality metrics (see Appendix \ref{appdx_sec:analysismethods} for detailed descriptions of each metric).While SI-SDR, PESQ, and STOI (shown as color) largely align, we note that this is not the case with the NMR-MOS model (shown as size).
    
       
    }
    
    \label{fig:audio_landscape}
\end{figure*}

 \begin{itemize}
\item Figure~\ref{fig:audio_landscape}: Audio Performance Landscape
 \item Figure~\ref{fig:age}: Age distribution.
\item Figure~\ref{fig:gender2}: Gender distribution.
\item Figure~\ref{fig:domains}: DOmains distribution.
 
   \item Figure~\ref{appdx_fig:WER_analysis}: WER distribution across languages for all ASR models.
   \item Figure~\ref{appdx_fig:CER_analysis}: CER distribution across languages for all ASR models.
      
\end{itemize}

\section{Framework} \label{appdx_sec:framework}

Please see Table \ref{tab:ideal_framework}.


\begin{table*}[h]
\centering
\resizebox{0.9\textwidth}{!}{%
\begin{tabular}{llll}
\toprule
Variation Category & Dimension & Metadata Representation & Example\\
\toprule
Region  & Dialect & ISO 639-3 & arz\\
Region &  Regional Accent & ISO 3166-2 & EG-C or EGY-C (Cairo) \\
Region &  `Rural-ness' & Categorical & Urban \\
\midrule
Stylistic  & Register & Description & Informal  \\
Stylistic  & Affect & Description  & Joyous \\
Stylistic  & Occasion & Description  & A family gathering \\
\midrule
Demographic & Age & Numeric & 40 \\
Demographic & Gender Expression & Categorical & M \\
Demographic & Ethnic Identity & Categorical & Egyptian \\
Demographic & Education & Categorical & Bachelors \\
Demographic & Socioeconomic & Categorical & Middle Class \\
Demographic & Language History & Description & L1 Arabic, L2 English speaker \\
Demographic & Speech Disfluencies & Description & Stutter present \\
\midrule
Recording & Channel & Description & Samsung Galaxy A16 smartphone \\
Recording & Environment & Description & Noisy; recording in a large room \\
Recording & Sampling Rate & Numeric & 16kHz \\
Recording & Encoding & Categorical & FLAC \\
\bottomrule

\end{tabular}%
}

\caption{An idealized framework for analyzing sources of variability in audio recordings for speech processing, alongside potential standardized ways these dimensions could be recorded in metadata. Open ended dimensions, such as style, lend themselves towards descriptive elements, while demographic elements might be better represented by speakers providing their own labels.}
\label{tab:ideal_framework}
\end{table*}

\section{Literature Review} \label{appdx_sec:lit_review}

\paragraph{Dataset Quality Analysis}
General-purpose dataset quality frameworks (e.g., \citealt{rosli2018evaluating}) are not tailored to the sources of variability that are specific to speech data, particularly recording and channel effects (Table~\ref{tab:ideal_framework}). Data documentation proposals such as \citet{bender2018data} provide a principled vocabulary for describing language variety, speakers, and speech situations, but they are primarily prospective standards and depend on creators to report these attributes consistently. In practice, many relevant factors are incompletely documented, and audio fidelity can be difficult to estimate at scale without matched clean references. Recent progress in non-intrusive (no-reference) quality estimation offers an alternative by providing model-based proxies intended to correlate with perceived quality \cite{kumar2023torchaudio,manocha22c_interspeech}. Therefore, we complement metadata with automated audio-quality measures computed directly from released training audio.

\paragraph{Speech Processing Methods}
Modern ASR systems predominantly build on transformer \cite{vaswani2017attention} and conformer \cite{gulati20_interspeech} backbones, spanning encoder-decoder architectures \cite{radford2023robust,omnilingual2025omnilingual}, CTC-based transformer encoders \cite{pratap2024scaling,omnilingual2025omnilingual}, and conformer encoders paired with transformer decoders \cite{barrault2023seamless}. The distinction between CTC-based and decoder-based systems is particularly relevant for dialectal settings: decoder-based models learn an explicit language model component that can bias outputs toward dominant training styles, whereas CTC systems operate via token-level alignment and may exhibit different failure modes \cite{pratap2024scaling}. This consideration becomes salient in large-scale multilingual training, where Arabic is often treated as a single language, potentially privileging MSA-like transcriptions, as observed in widely used models such as Whisper \cite{radford2023robust} and MMS \cite{pratap2024scaling}. Therefore, our evaluation compares model families with different decoding biases to better understand performance on dialectal Arabic.

\section{Mapping and Standardization Pipeline}\label{appdx_sec:mapping}
We begin this appendix with descriptions of all the datasets, and also provide a summary table of the total hours of these datasets broken into dialect (Table \ref{tab:arabic_dialects_main}).


\begin{table*}[!h]
\centering
\resizebox{0.95\textwidth}{!}{%

\begin{tabular}{lllrrrrrrrr}
\toprule
\multirow{2}{*}{\textbf{Dialect}} & \multirow{2}{*}{\textbf{Primary Countries}} & \multirow{2}{*}{\textbf{ISO}} &
\multicolumn{2}{c}{\textbf{TRAIN}} & &
\multicolumn{2}{c}{\textbf{DEV}} & &
\multicolumn{2}{c}{\textbf{TEST}} \\ \cmidrule{4-5} \cmidrule{7-8} \cmidrule{10-11}
& & &\textbf{ Dur (H)} & \textbf{Uttr} & &
 \textbf{Dur (H)} & \textbf{Uttr} & &
 \textbf{Dur (H)} & \textbf{Uttr} \\ \toprule
Levantine & JOR, LBN, PSE, SYR & apc & 23.2 & 18,581 & & 1.0 & 974 & & 9.5 & 7,415 \\
Khaleeji  & KWT, QAT, ARE, SAU & afb & 96.2 & 62,235 & & 1.1 & 922 & & 3.0 & 2,204 \\
MSA  & - & arb & 3951.6 & 2,217,449 & & 36.1 & 26,484 & & 36.6 & 27,201 \\
\midrule
Algerian  & DZA & arq & 0.5 & 600 & & 1.0 & 844 & & 1.0 & 921 \\
Egyptian  & EGY & arz & 427.2 & 22,872 & & 7.4 & 5,554 & & 7.8 & 4,893 \\
Hassaniya & MRT, MLI, ESH & mey & -- & -- & & 1.0 & 953 & & 0.9 & 953 \\ 
Libyan  & LBY & ayl & 0.0 & 37 & & -- & -- & & -- & -- \\
Moroccan  & MAR, ESH & ary & 39.5 & 31,266 & & 8.5 & 6,985 & & 8.9 & 6,559 \\
Sudanese  & SUD &  apd & 4.3 & 3,933 & & -- & -- & & 0.1 & 127 \\
Tunisian  & TUN & aeb & 61.7 & 21,139 & & 0.0 & 36 & & 2.3 & 2,060 \\


Hijazi & SAU & acw & 40.2 & 34,833 & & 0.6 & 528 & & 1.1 & 809 \\
Najdi   & SAU & ars & 117 & 90,657 & & 3.3 & 2,250 & & 2.1 & 1,704 \\
Mesopotamian  & IRQ & acm & 32 & 20,848 & & 0.4 & 446 & & 2.2 & 1,511 \\
North Mesopotamian  & IRQ & ayp & 4.7 & 3,032 & & -- & -- & & 0.7 & 414 \\
Sanaani & YEM & ayn & 0.70 & 1,115 & & -- & -- & &  -- & -- \\
Ta'izzi-Adeni   &  YEM &acq & 0.8 & 1,224 & & -- & -- & & -- & -- \\
\midrule
Total & - & - & 4799.6 & 2,529,821 & & 60.4 & 45,976 & & 76.2 & 56,771 \\

\midrule
Unspecified & unk & unk & 1,371.4 & 634,411 & & 11.8 & 14,278 & & 32.80 & 25,753 \\
\bottomrule
\end{tabular}%
}
\caption{Dialect-wise breakdown of transcribed Arabic audio by train/dev/test splits (hours and utterances). Asterisks ($^\ast$) indicate total hours after splitting datasets without canonical splits of train, dev, and test.  We do not find any datasets corresponding to the following ISO codes:  Algerian Saharan Arabic (aao), Tajiki Arabic (abh), Baharna Arabic (abv), Omani (acx),  Cypriot (acy), Dhofari (adf), Saidi (aec), Uzbecki (auz), Eastern Egyptian Bedawi Arabic  (avl), Hadrami (ayh), Sudanese Creole Arabic (pga), Chadian Arabic (shu), and Shihhi Arabic (ssh). While we did identify small portions of Libyan (ayl), we did not have enough to constitute a credible piece of our final benchmark.
}
\label{tab:arabic_dialects_main}
\end{table*}

\subsubsection*{Monolingual Transcribed Datasets}

\paragraph{ASR-EgArbCSC}\cite{magicdata_no_date_egyconv}: 5.5 hours of conversational Egyptian speech. While demographic information is provided (age, gender, city), and information about the conversational topic is also provided, there is little other information regarding the provenance and collection strategy used for this dataset.

\paragraph{ASR-YeArCSC}\cite{magicdata_no_date_yemconv}: 10.42 hours of conversational Yemeni speech. While demographic information is provided (age, gender, city), and information about the conversational topic is also provided, there is little other information regarding the provenance and collection strategy used for this dataset. This is somewhat problematic, as language information may need to be inferred based on the city information.

\paragraph{CALLHOME}\cite{LDC97S45_callhome_egypt}: This 60 hours Egpytian Arabic corpus consists of unscripted telephone calls. A selection of each call is provided with a transcript, meaning that far fewer than the full total amount of time is available for use in training ASR systems (14.3 /  3.6 / 1.7 hr.). Additional details are provided about each recording, for instance for elderly speakers, gender of speakers, and any challenging conditions, however, little additional demographic information is provided for recipients of the calls.

\paragraph{Casablanca}\cite{talafha2024casablanca}: This dataset consists of 48 hours dialect Arabic of which a validation and test split of 8 hours each has been released. Eight country-level dialects are included covering: Algeria, Egypt, Jordan, Morocco, Mauritania, Palestine,  UAE, and Yemen. Audio was collected from YouTube, with manual transcription and validation. Gender splits vary widely between country ranging from over 92\% male for Palestine to 57\% male for Morocco. 

\paragraph{Common Voice}\cite{ardila2019common}:
The most recent version (as of writing) of Common Voice, 21, consists of 92 hours of validated read speech. Sentences are read by volunteers and recorded on their own devices, for instance a laptop mic, with the sentences themselves submitted and verified by volunteers. The sentences selected are MSA, although Common Voice does support dialects of other languages (for Cantonese and Minnan for Chinese on top of Mandarin).

\paragraph{FLEURS}\cite{conneau2023fleurs}: 
This smaller dataset, consisting of 8 hours of MSA. To enhances multilingual speech research, the FLEURS dataset was proposed using aligned text for all languages, originally sourced from English Wikipedia as part of the FLORES-101 project~\cite{goyal2022flores}. FLEURS takes these sentences and recruited native speakers to record each sentence. In the case of the Arabic split of the dataset,  Egyptian speakers, with varying degrees of accent, provided the voices.

\paragraph{GALE} \cite{LDC2013S07_GALE_p2_conv_1}\footnote{For simplicity this is the first of the GALE Arabic speech series}: GALE phases 2-4 produced both conversational (578 hr.) and broadcast news (632 hr.) datasets for Arabic speech. While mainly MSA, a significant amount is in DA, which has lead to additional work to identify, extract, and use the DA segments for ASR systems ~\cite{soltau2011modern, alsharhan2020investigating}. In annotating parts of GALE Phase 3 \cite{alsharhan2020investigating} identified 44 hours of DA, mainly consisting of Levantine dialect. Gender imbalance also persists between dialects, with the work of \cite{alsharhan2020investigating} indicating, gender ratios of at most 41\% female speakers with Levantine dialect, and as little as 5\% with Iraqi,  of the amount they annotate.

\paragraph{Gulf Conversational Speech (Appen)}\cite{LDC2006S43_gulf_speech}: This conversational speech datasets consists of roughly 47 hours of Gulf dialect speech. The  associated transcripts provide single references with full diacritization.

\paragraph{Iraqi Conversational Speech (Appen)}\cite{LDC2006S45_iraqi_speech}: A 50 hour telephone speech dataset consisting of spontaneous conversational speech. Like the other Appen dataset, this is a 8khz recording with single reference but fully diacritized transcripts.

\paragraph{L2-KSU}\cite{LDC2024S11_l2_ksu}: This small (6 hour) dataset offers a unique selection of L1 and L2 Arabic speakers recording read MSA sentences. The majority of the L2 speakers are from African language backgrounds, while L1 Arabic speakers reflect a mixture of Egyptian, Gulf, and Levantine dialect backgrounds.

\paragraph{MASC}\cite{al2023masc}
The Massive Arabic Speech Corpus is a 1000 hour multigenre and multidialectal dataset sourced from Arabic YouTube channels with manually uploaded transcripts. A majority (569 hrs) of the language being MSA, and the top 5 represented dialects being: Syrian (197 hrs), Egyptian (120 hrs), Jordanian (39 hours), Saudi (31 hours) and Lebanese (23 hrs). The audio was lightly filtered with manual checking to ensure the captions were in Arabic, however, alignment issues may remain. A large majority of the speakers were identified as male accounting for 74\% of the main speakers (excluding mixed gender segments) and accounting for 79\% of the total speaking time. Text preprocessing involved using the Maha library to perform Alef normalization, including splitting Lam Alef to separate Lam and Alef characters, and normalize Teh Marbuta to Heh.  The dataset is provided in 16khz, 16-bitdepth, 1-channel recordings.

\paragraph{MGB-2}\cite{ali2016mgb}: The multigenre broadcast corpus 2 consists of over 1,200 hours of mainly MSA audio sourced from 19 different Al Jazeera programs covering a wide range of program formats. The audio is mainly political coverage, with 24\% falling into other topics which the authors indicate include society, economics, media, law, and science. The original paper does not give a very precise estimate of the amount of dialectal language used (only that it accounts for no more than 30\%), however \cite{mubarak2021qasr} performed an analysis of the test set, and indicates 78\% MSA with 22\% dialectal. Gender imbalance is present in the dataset, with \cite{mubarak2021qasr} indicating a ratio of 78\% (male) to 11\% (female) speakers in the MGB-2 test set. These ratios are not known for the training split of the MGB-2. While there is no overlap in segments from the training to the evaluation splits, no speaker linking has been performed to confirm that there is not a speaker overlap. Surface normalization of the text is applied. 

\paragraph{MGB-3}\cite{ali2017speech}: 15.4 (4.6/4.8/6.0) hour Egyptian Arabic dataset sourced from YouTube videos covering a number of different genres (balanced across seven broad categories such as ``sports'' and ``family''). The authors note 1.4 hours include overlapping speakers with the rest of the audio single speaker recordings.  Four different transcriptions are available for each recording to account for the lack of standardized orthography. The authors do not provide demographic information about this dataset. Surface normalization is applied for alef, yah, and hah.

\paragraph{MGB-5}\cite{ali2019mgb}: 13 (10.2/1.3/1.4) hour Moroccan Arabic dataset sourced from YouTube videos and following a similar strategy as MGB-3. This includes a balance of different genres, multi reference with four different annotations, and surface text normalization. The largest departure from MGB-3 is in the way that the dataset has been split, with significantly smaller development and testing sets in comparison to MGB-3. No demographic information is provided.

\paragraph{SADA}\cite{alharbi2024sada}: This corpus of 668 hours of Saudi dialect speech (of which 437.6 hours have transcribed labels), represents one the largest transcribed dialectal datasets. There is a breadth of coverage in terms of Saudi dialect and genre of program (Comedy, Drama, Cooking etc.). Surface normalization was applied to the transcripts, and standard spelling is used for MSA, however, it is unclear how annotators agreed on transcription standards for the dialectal Arabic. While most of the dataset corresponds to ISO-639-3 level language codes (e.g. Najdi Arabic), there is ambiguity with how they define Maghrebi Arabic (which could be one of a number of North African Arabic languages), as well as their use of Northern (\textit{Shamali}) and Southern (\textit{Janubi}).

\paragraph{QASR}\cite{mubarak2021qasr}: 2041 hour dataset mainly containing MSA from multigenre Al Jazeera recordings. The majority of language is expected to be MSA, however, dialectal Arabic is present.
A small sample of 6,000 utterances were identified as containing code-switching between MSA, English, and/or French. A split of 4,000 segments were annotated for Speaker and Dialect ID purposes.

\paragraph{Quran Speech to Text (OpenSLR)}\cite{slr132}: 
Recordings of verses of the Quran taken from \url{https://quran.ksu.edu.sa} and resampled to 16khz. As these are recited verses, they are potentially closer to sung utterances than read utterances, and may provide particular distinct domain of audio.

\paragraph{Tunisian MSA (OpenSLR)}\cite{slr46}: This 11.2 hour dataset consists of read and prompted utterances, from Tunisian speakers in MSA. No additional demographic information is provided, nor does it contain any documentation of methodology for data collection.

\subsubsection*{Codemixed Datasets}
Codemixing refers to the use of two different languages within a single utterance~\cite{albirini2011sociolinguistic}, and is quite common in real world dialectal Arabic conversations, with the language mixed in dependent on the country and context.



\paragraph{ArzEn}\cite{hamed2020arzen}: A 12 hour dataset consisting of spontaneous speech from interviews with Egyptian university students and teaching assistants (with the addition of one university employee). The focus of this dataset is on English and Egyptian Arabic codemixing (one of the two main modes of codemixing in Egyptian society, the other being MSA and Egyptian Arabic), with the interview style setting supportive of this natural speaking style. The vast majority of utterances (89\%) consist of Egyptian Arabic with English inserted.

\paragraph{Mixat}\cite{mixat}: 
This 14.9 hour dataset leverages podcast audio of hosts that naturally codemix between Emirati (a Gulf Arabic dialect) and English in their recordings. Unlike Arzen, which consists mainly of codemixed sentences, Mixat only contains a portion (36\%) which contain codemixing. The majority is monolingual Emirati Arabic.

\paragraph{Saudilang Code-Switch Corpus (SCC)}\cite{SCC2025}: 5 hour dataset sourced from Thmanyah podcast. The datasets consists of English and Arabic code-switch dialogues designed primarily as an out-of-domain test set.

\paragraph{ZAEBUC-Spoken}\cite{hamed-etal-2024-zaebuc}: Using a fairly novel Zoom-based collection method, this dataset consists of brainstorming discussion between students. The study used a mix of English speaking facilitators (from a wide range of accents), as well as Arabic speaking facilitators (primarily speaking MSA, but occasionally MSA with Egyptian codemixing), and the students spoke in a mix of English and Gulf Arabic, as well as MSA and Gulf Arabic. The majority of the utterances are either monolingual English or Arabic, with codemixing Eng-Arabic accounting for only 16\% of the total utterances. For MSA-Dialectal codemixing, only a portion of the Arabic utterances were annotated using a five-level scale of dialectness. The majority of these utterances (58\%) consisted of either pure MSA or imperfect MSA, without dialectal markers.

\subsubsection*{TTS-oriented datasets }
The following speech datasets indicate suggested use for training TTS systems.

\paragraph{ASC}\cite{halabi2016modern}: 
Single speaker speech corpus 3.7 hour datasets with aligned diacritized texts for the purpose of training MSA TTS systems. Detailed diacritization and phonetic information is provided.

\paragraph{ClArTTS}\cite{kulkarni2023clartts}: 
In contrast to many of the other datasets, this 12 hour dataset uses classical Arabic . Recorded at 44.1kHz.

\paragraph{ArVoice}\cite{toyin2025arvoicemultispeakerdatasetarabic}:

83.52 hours of which 73.5 is synthetic audio (11 voices) and 10 hours of human speakers (7 voices).
The origin of the text is Tashkeela, Khaleej, and a modified version of the texts used in ASC\cite{halabi2016modern}. All texts are written in MSA.

\paragraph{Iraqi TTS}\cite{kharrufa_2024_11170567}: 
With 3.7 hours of MSA and 1 hour of dialectal Arabic speech, this dataset is designed for training TTS systems that can handle a mix of MSA and dialectal speech. Lack of description of the speakers, as well as lack of precision in describing the dialect limit its utility.




\subsubsection*{Emotion Recognition Datasets}
Some datasets have additional emotion labeling for emotion recognition models.

\paragraph{Kasdi-Merbah University Emotional Database of Arabic Speech}\cite{LDC2023S10_kasdi}:
This is a small, two hour, dataset aimed at emotion recognition. Notably for the small size of the dataset a large number of speakers are included including 254 female, 246 male speakers, each reading 10 sentences. The audio was recorded at 44 kHz.  

\paragraph{KSUEmotions}\cite{LDC2017S12_KSUEmotions}: 5 hours of emotion-labeled speech. The speech is read newswire, however, it is unclear how the choice of news text paired with somewhat artificially selected emotions impact the real world performance of emotion classifier trained on this dataset.

\section{Dataset Analysis}\label{appdx_sec:analysismethods}

\paragraph{Dialect Coverage}
Across the datasets we find a large difference in the quantity of audio for each dialect, with limited amounts for Algerian (arq), Yemeni dialects (acq, ayn, and ayh), Sudanese (apd), Libyan (ayl), Tunisian (aeb), and Hassaniya (mey) (see Table \ref{tab:arabic_dialects_main}).

\paragraph{Age}

\begin{figure}[ht]
    \centering
    \includegraphics[width=\columnwidth]{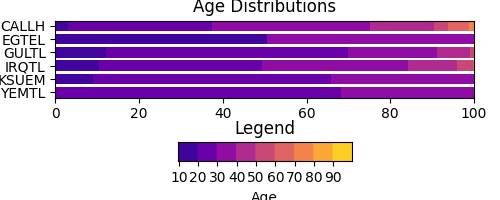}
    \caption{Percentage distribution of age ranges in Datasets with age information of speakers. Speakers binned to 10 year age intervals.}
    \label{fig:age}
\end{figure}

Of the datasets, only six provide age demographics for speakers, which we provide in Figure ~\ref{fig:age}. Of the datasets which provide age information, there appears to be a relative lack of older speakers represented in the datasets, with the vast majority of speakers under 40 years old.

\paragraph{Gender}

\begin{figure}[ht]
    \centering
    \includegraphics[width=\columnwidth]{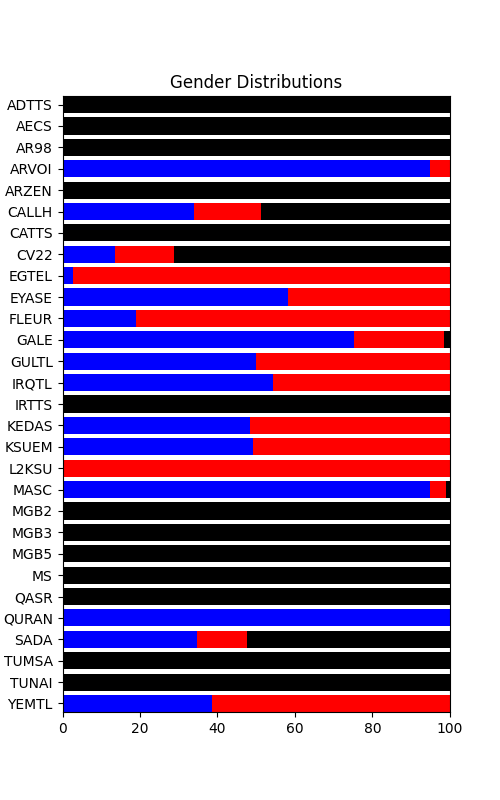}
    \caption{Percentage of Male (blue), Female (red) and unknown (black) speakers where datasets report gender.}
    \label{fig:gender2}
\end{figure}

A common challenge in speech processing dataset curation is ensuring a relatively equal gender distribution of speakers. While normative expectations to obtain balanced gender ratios may not guarantee fair gender performance in general (see \cite{garnerinInvestigatingImpactGender2021}), few papers included gender information (16 out of 28), and those that did often indicated wildly imbalanced ratios that may be dependent on the collected country ~\cite{talafha2024casablanca}.
Existing benchmarks rarely provide gender breakdown of performance, further hindering development of equitable speech processing systems. Figure ~\ref{fig:gender2} provides breakdown of the duration from a given gender, with quite a few are extremely unbalanced, with seven reporting over 80\% of the duration belonging to just one gender.
Fig. ~\ref{fig:gender2} showns an overview of the gender distributions.


\paragraph{Domain}
\begin{figure*}[!ht]
    \centering
    \includegraphics[width=\textwidth]{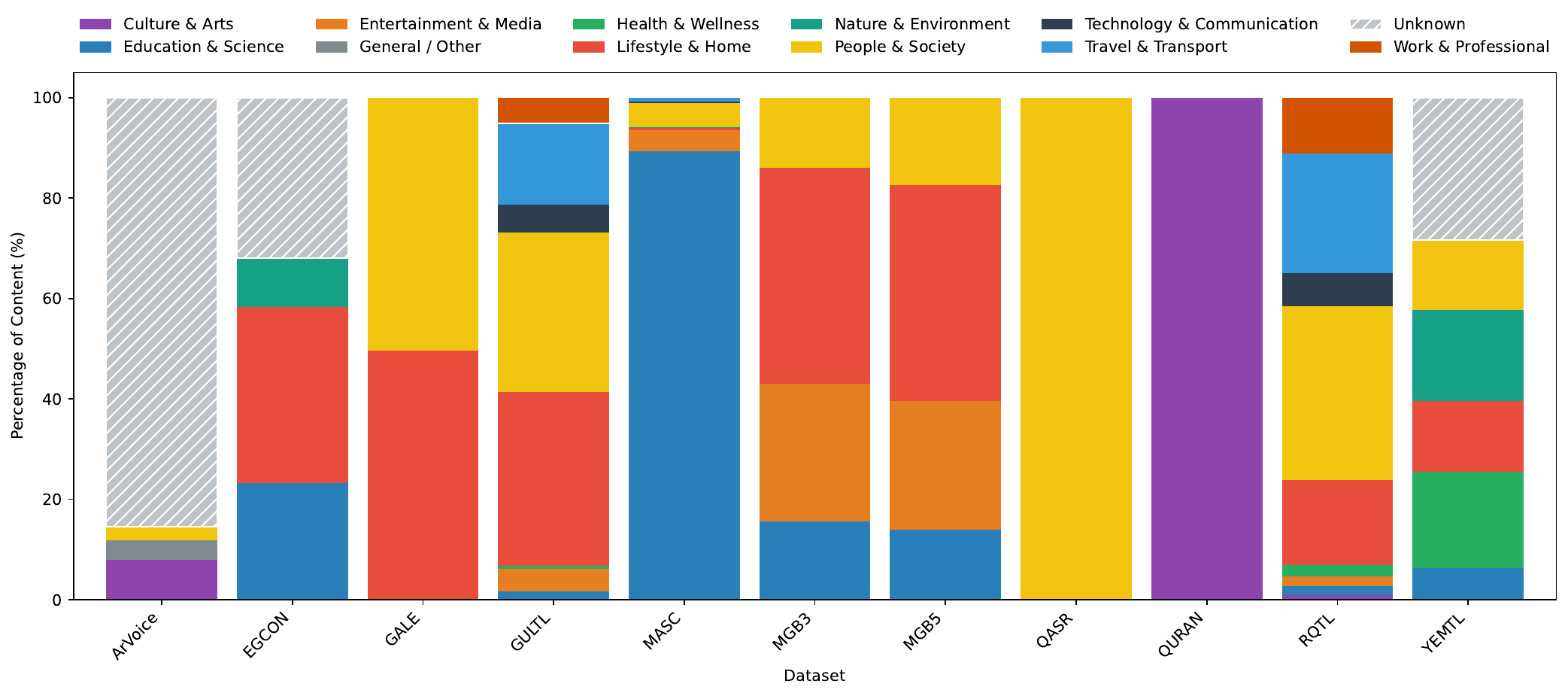}
    
    \caption[]{
        Per-dataset, per-domain breakdown of utterance counts and total hours. Dataset names are shown only on the first row of each block. We omit dataset which do not provide any information about domain (11 out of the 28). 
    }
    
    \label{fig:domains}
\end{figure*}

We provide a full breakdown of domain per dataset in Fig. \ref{fig:domains}. We find that the distribution of domains is not particularly diverse, with People \& Society accounting for the vast majority of utterances, however, utterances with unknown domain (either a generic label, ``misc,'' or not provided) account for the largest amount of total duration (2625 hours).

\subsection{Dialectness}
\label{sec:dialectalness_details}

\begin{figure}[ht]
    \centering
    \includegraphics[width=0.9\columnwidth]{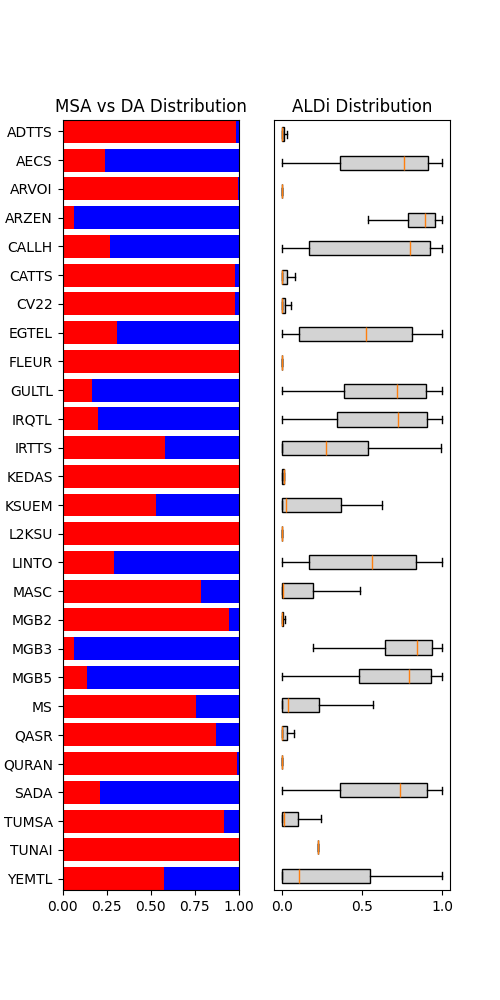}
    \caption{(Left) Percentage of MSA (red) and DA (blue), based on predictions from our binary classifier. (Right) ALDi box plots, a higher ALDi score implies more dialectness. }
    \label{fig:msa_da}
\end{figure}

\paragraph{Binary MSA vs DA classifier}:
We use a finetuned version of MARBERTs~\cite{abdul2021arbert}, finetuned on an in-house text dataset. 

\paragraph{ALDi regression model}: Also a finetuned MARBERT~\cite{abdul2021arbert}, trained on the Arabic Online Commentary (AOC)-ALDi dataset, which is augmented version of the AOC dataset~\cite{zaidan2011arabic}. In contrast to the binary model, which have a classification head, ALDi uses a regression approach to provide an estimate between 0 and 1 of the level of dialectness. The original ALDi-AOC dataset uses the following bins: [0, 0.11[; [0.11, 0.44[; [0.44, 0.77[; [0.77, 1[ for MSA, little DA, mixed, and mostly DA respectively.

A limitation of textual approaches, however, is that they are not able to identify sentences, where the dialectal and MSA sentence would be written the same way without diacritics.

\paragraph{Detailed Charts}

For ALDi we provide violin plots showing the distribution of ALDi scores from 0 (MSA) to 1 (purely dialect). Quartiles are shown in light black.

\includegraphics[width=0.4\linewidth]{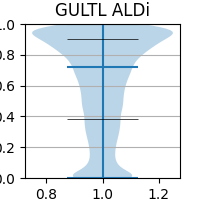}
\includegraphics[width=0.4\linewidth]{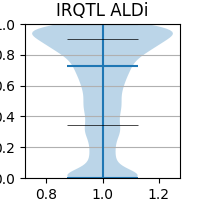}
\includegraphics[width=0.4\linewidth]{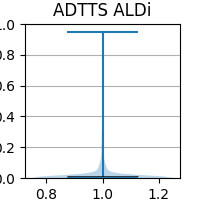}
\includegraphics[width=0.4\linewidth]{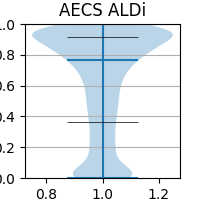}
\includegraphics[width=0.4\linewidth]{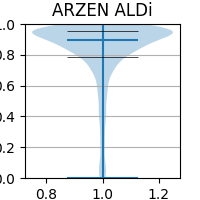}
\includegraphics[width=0.4\linewidth]{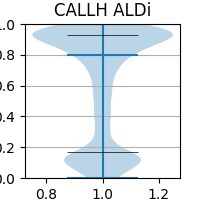}
\includegraphics[width=0.4\linewidth]{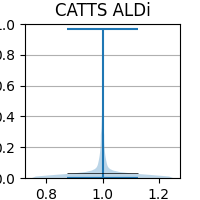}
\includegraphics[width=0.4\linewidth]{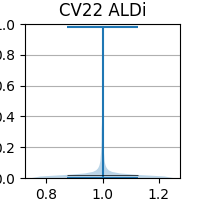}
\includegraphics[width=0.4\linewidth]{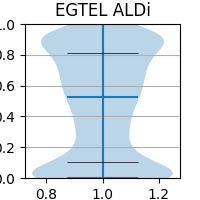}
\includegraphics[width=0.4\linewidth]{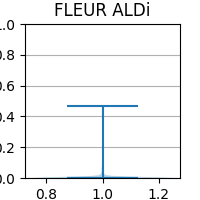}
\includegraphics[width=0.4\linewidth]{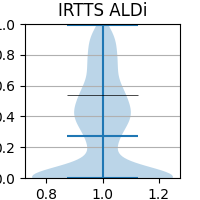}
\includegraphics[width=0.4\linewidth]{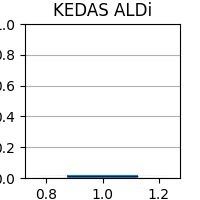}
\includegraphics[width=0.4\linewidth]{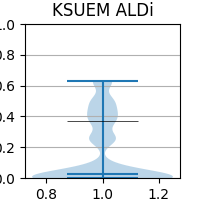}
\includegraphics[width=0.4\linewidth]{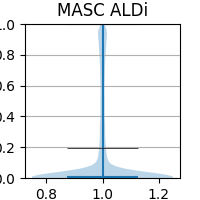}
\includegraphics[width=0.4\linewidth]{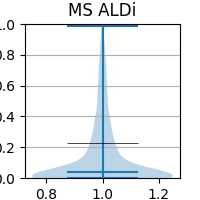}
\includegraphics[width=0.4\linewidth]{img/mediaspeech-ar_aldi.png}
\includegraphics[width=0.4\linewidth]{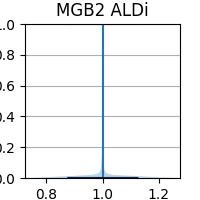}
\includegraphics[width=0.4\linewidth]{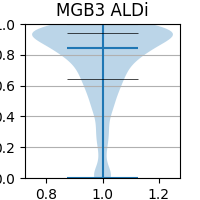}
\includegraphics[width=0.4\linewidth]{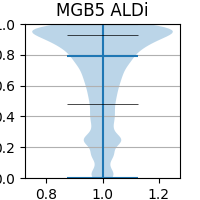}
\includegraphics[width=0.4\linewidth]{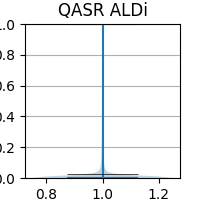}
\includegraphics[width=0.4\linewidth]{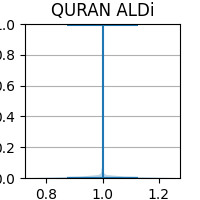}
\includegraphics[width=0.4\linewidth]{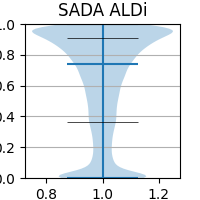}
\includegraphics[width=0.4\linewidth]{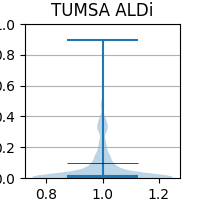}
\includegraphics[width=0.4\linewidth]{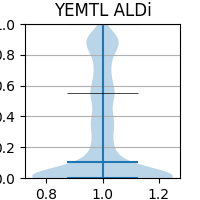}

\subsection{SQUIM}
\label{sec:squim}

PESQ can be thought of as a prediction of subjective human perception of audio quality (albeit measured with an objective model), and includes cognitive and auditory modeling to predict a mean opinion score in the range [1,4.5]~\cite{rix2001perceptual}. SI-SDR, is an updated signal-to-noise metric that accounts for issues in the signal-to-noise formula caused by the scaling of the estimated audio~\cite{le2019sdr}, the updated SI-SDR formula accounts for this to provide a metric that is independent of amplitude scaling of the audio. Like signal-to-noise, the metric is measured in decibels, and represent how many orders of magnitude the target audio is compared to any distortions. Finally, STOI measures intelligibility of audio based on 382ms audio chunks, and measures the correlation between chunks of clean and noisy audio, with the higher alignment indicating better intelligibility of the audio file~\cite{taal2011algorithm}.

TorchAudio-Squim uses a transformer based model to simulate these metrics without the need for isolated clean audio. While these objective measures of audio quality may not correlate well with human judgment~\cite{manocha22_interspeech}, they may still have an impact on model performance for models that are highly sensitive to domain and channel information, and by providing this information, we aim to set the stage for future evaluation of these factors. As an alternative, we also use Squim's subjective model, which is based on the non-matching reference approach in ~\cite{manocha22c_interspeech}, this model provides a prediction correlated with human mean opinion score (MOS) ratings. We randomly sample 5 seconds of clean audio from the DAPS dataset~\cite{mysore2014can} to provide our non-matching reference.

As PESQ and the Squim-Subjective metric are aligned to mean opinion scores, they are the most interpretable of the metrics, higher scores should thus align to human perception of audio quality. On the other hand, SI-SDR, which reflects the relative ratio of target audio to distortion on a decibel scale; and STOI with its 0 to 1 rating of intelligbility, both lack easily interpretable meaning. However, they still provide a valuable metric when compared to other utterances on the same scale as a way to characterize the noisiness of a particular dataset.

\subsection{Noise and intelligibility.}

We next analyze predicted recording conditions using our audio-quality proxies. Consistent with the recording setups typically used for TTS corpora (Table~\ref{tab:squim}), ADTTS, CATTS, and IRTTS exhibit high predicted quality under the objective proxies (all mean predicted PESQ $>3$ and mean predicted SI-SDR $>20$). Several other read-speech datasets, including ARVOI, TUNAI, and TUMSA, show similarly strong scores. Within broadcast-domain recordings, QASR scores well by these proxies and appears higher quality than the similarly situated MGB2 dataset as well as the other large-scale MSA dataset, MASC.

In contrast, mobile and telephone recordings tend to score lower on these proxies, with several datasets exhibiting mean predicted PESQ $\leq 2$ and mean predicted SI-SDR $<10$, including GULTL, IRQTL, and CALLH. Other notable datasets include SADA and QURAN, which have mean predicted PESQ $<2$, mean predicted SI-SDR $<10$, and mean predicted STOI around 0.8.

Predicted NMR-MOS yields a different ordering from the objective proxies: CALLH and ARVOI are both assigned relatively high predicted MOS (3.7), despite representing opposite ends of the predicted objective-quality spectrum (PESQ: ARVOI 3.8 vs.\ CALLH 1.6). \S\ref{appdx_sec:analysismethods} provides detailed figures for PESQ, SI-SDR, and STOI for each dataset. Therefore, rather than collapsing quality into a single score, our benchmark/adaptation protocol retains dataset provenance and reports results in the context of these complementary audio-condition signals.

\begin{table}[!ht]
\centering
\resizebox{0.5\textwidth}{!}{%
\begin{tabular}{lcccc}
\toprule
Dataset & PESQ $\uparrow$ & SI-SDR (dB) $\uparrow$ & STOI $\uparrow$ & NMR-MOS $\uparrow$ \\
\toprule
ARVOI & 3.8 (0.3) & \high{25.7} (3.1) & \high{1.0} (0.0) & \high{3.7} (0.9) \\
GULTL & 2.0 (0.6) & 8.5 (9.4) & 0.9 (0.1) & 3.3 (0.8) \\
IRQTL & 2.0 (0.6) & 9.9 (9.4) & 0.9 (0.1) & 3.3 (0.9) \\
ADTTS & 3.8 (0.3) & \high{25.0} (3.2) & \high{1.0} (0.0) & 3.4 (0.9) \\
AECS & 2.8( 0.8) & 17.9 (6.3) & \high{1.0} (0.0) & 3.1 (0.9) \\
ARZEN & 3.1 (0.8) & 18.6 (8.0) & \high{1.0} (0.1) & \high{3.5} (0.8) \\
CALLH & \low{1.6} (0.6) & \low{4.2} (9.3) & \low{0.8} (0.2) & \high{3.7} (0.6) \\
CATTS & 3.3 (0.5) & 23.1 (3.2) & \high{1.0} (0.0) & 3.1 (0.6) \\
CV22 & 2.7 (0.8) & 18.9 (8.1) & 0.9 (0.1) & 3.4 (0.7) \\
EGTEL & 2.1 (0.9) & \low{6.2} (16.4) & \low{0.8} (0.2) & \low{3.0} (1.1) \\
FLEUR & 2.7 (0.7) & 20.4 (6.2) & \high{1.0} (0.0) & 3.2 (0.9) \\
IRTTS & \high{4.1} (0.2) & \high{25.8} (2.0) & \high{1.0} (0.0) & \high{3.5} (0.8) \\
KEDAS & 2.9 (0.9) & 14.9 (11.8) & \high{1.0} (0.0) & 3.4 (0.7) \\
KSUEM & 2.5 (0.6) & 19.3 (6.0) & \high{1.0} (0.1) & 3.3 (0.8) \\
L2KSU & 3.0 (0.7) & 20.4 (5.8) & \high{1.0} (0.0) & 3.2 (0.9) \\
LINTO & 2.3 (1.0) & 13.5 (10.9) & 0.9 (0.1) & 3.4 (0.7) \\
MASC & 2.3 (0.9) & 13.8 (10.7) & 0.9 (0.1) & 3.3 (0.8) \\
MS & 2.6 (0.7) & 16.6 (6.1) & \high{1.0} (0.0) & \low{2.9} (0.8) \\
MGB2 & 2.9 (0.7) & 19.4 (7.0) & \high{1.0} (0.1) & 3.3 (0.9) \\
MGB3 & 2.6 (0.8) & 16.7 (7.7) & 0.9 (0.1) & 3.4 (1.0) \\
MGB5 & 2.1 (0.8) & 11.8 (10.8) & 0.9 (0.2) & 3.3 (0.9) \\
QASR & 3.4 (0.6) & 22.4 (5.3) & \high{1.0} (0.0) & 3.3 (0.6) \\
QURAN & 1.8 (0.8) & 7.6 (9.7) & \low{0.8} (0.1) & 3.4 (1.1) \\
SADA & 1.8 (0.6) & 9.5 (10.1) & \low{0.8} (0.2) & 3.2 (0.9) \\
TUNAI & \high{4.0} (0.2) & \high{26.3} (2.5) & \high{1.0} (0.0) & \high{3.5} (0.6) \\
TUMSA & 3.5 (0.4) & \high{24.7} (3.2) & \high{1.0} (0.0) & \low{3.0} (0.7) \\
YEMTL & 2.3 (0.9) & 13.9 (16.4) & \low{0.8} (0.2) & \high{3.6} (0.6) \\
\bottomrule

\end{tabular}%
}
\caption{Audio metrics overview. Values are reported as Mean (Standard Deviation). For bounded metrics the ranges are PESQ [1, 4.5], STOI [0, 1], SUB [1, 5]; SI-SDR scores are in decibels (dB). Notable low scores are highlighted in \low{red}, and higher quality scores in \high{dark green}. }
\label{tab:squim}\end{table}

\subsection{SQUIM PESQ}

\includegraphics[width=0.4\linewidth]{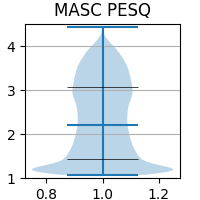}
\includegraphics[width=0.4\linewidth]{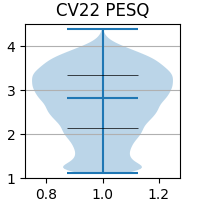}
\includegraphics[width=0.4\linewidth]{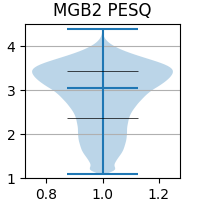}
\includegraphics[width=0.4\linewidth]{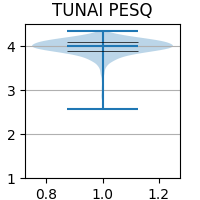}
\includegraphics[width=0.4\linewidth]{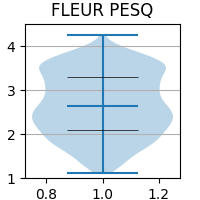}
\includegraphics[width=0.4\linewidth]{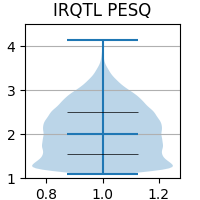}
\includegraphics[width=0.4\linewidth]{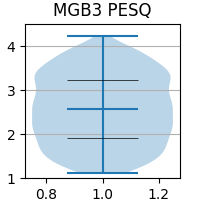}
\includegraphics[width=0.4\linewidth]{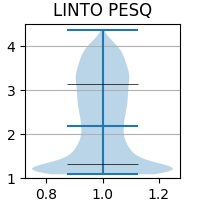}
\includegraphics[width=0.4\linewidth]{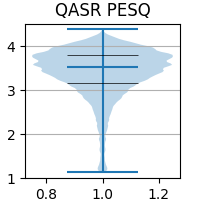}
\includegraphics[width=0.4\linewidth]{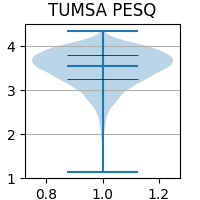}
\includegraphics[width=0.4\linewidth]{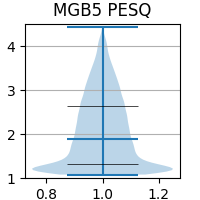}
\includegraphics[width=0.4\linewidth]{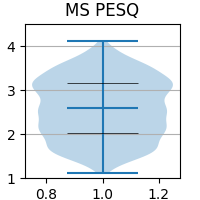}
\includegraphics[width=0.4\linewidth]{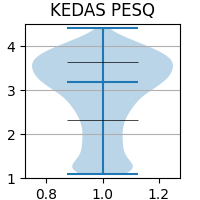}
\includegraphics[width=0.4\linewidth]{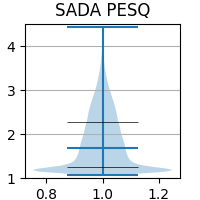}
\includegraphics[width=0.4\linewidth]{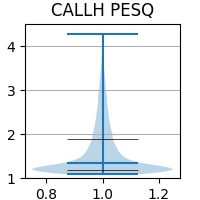}
\includegraphics[width=0.4\linewidth]{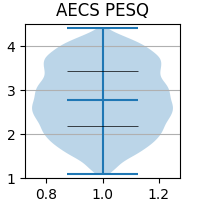}
\includegraphics[width=0.4\linewidth]{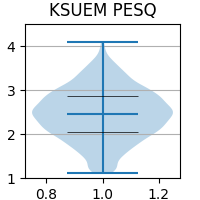}
\includegraphics[width=0.4\linewidth]{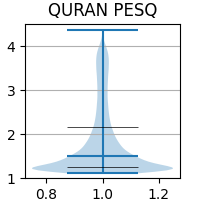}
\includegraphics[width=0.4\linewidth]{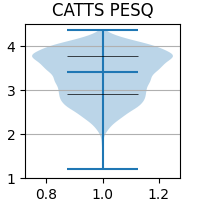}
\includegraphics[width=0.4\linewidth]{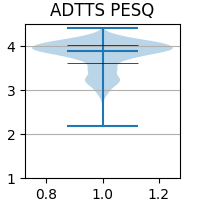}
\includegraphics[width=0.4\linewidth]{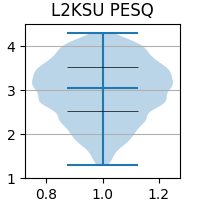}
\includegraphics[width=0.4\linewidth]{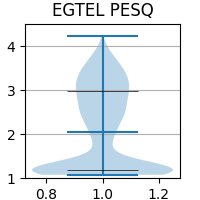}
\includegraphics[width=0.4\linewidth]{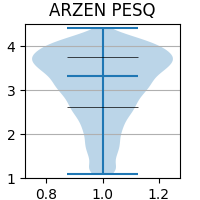}
\includegraphics[width=0.4\linewidth]{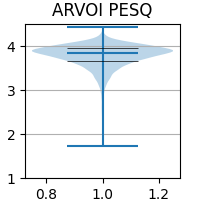}
\includegraphics[width=0.4\linewidth]{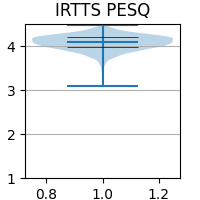}
\includegraphics[width=0.4\linewidth]{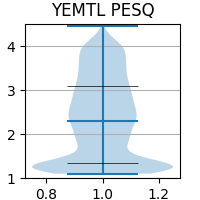}
\includegraphics[width=0.4\linewidth]{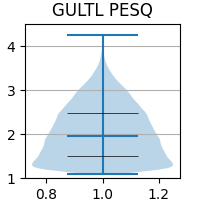}

\subsection{SQUIM SI-SDR}
\includegraphics[width=0.4\linewidth]{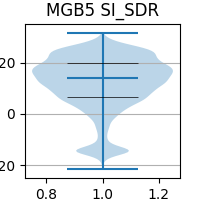}
\includegraphics[width=0.4\linewidth]{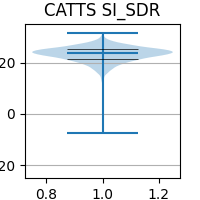}
\includegraphics[width=0.4\linewidth]{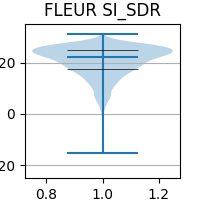}
\includegraphics[width=0.4\linewidth]{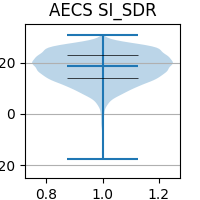}
\includegraphics[width=0.4\linewidth]{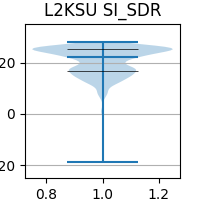}
\includegraphics[width=0.4\linewidth]{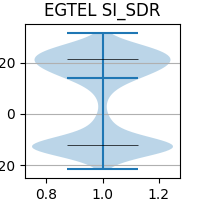}
\includegraphics[width=0.4\linewidth]{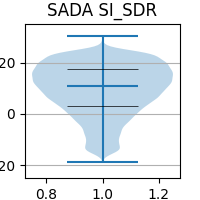}
\includegraphics[width=0.4\linewidth]{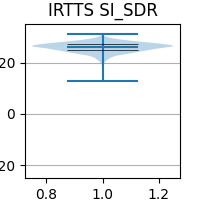}
\includegraphics[width=0.4\linewidth]{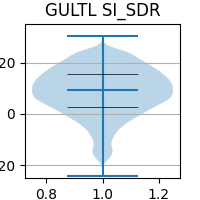}
\includegraphics[width=0.4\linewidth]{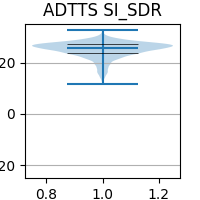}
\includegraphics[width=0.4\linewidth]{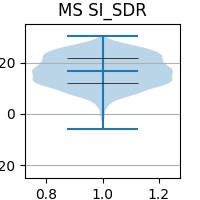}
\includegraphics[width=0.4\linewidth]{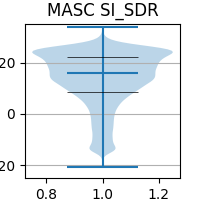}
\includegraphics[width=0.4\linewidth]{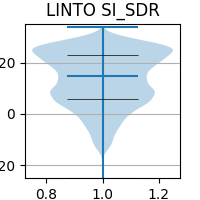}
\includegraphics[width=0.4\linewidth]{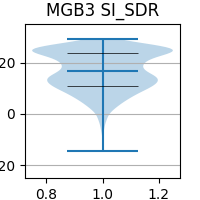}
\includegraphics[width=0.4\linewidth]{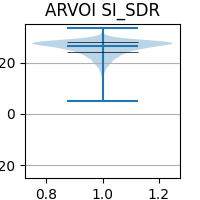}
\includegraphics[width=0.4\linewidth]{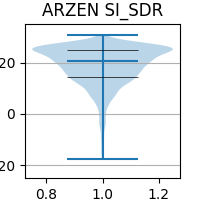}
\includegraphics[width=0.4\linewidth]{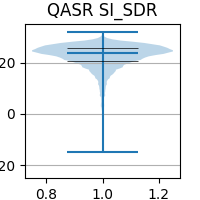}
\includegraphics[width=0.4\linewidth]{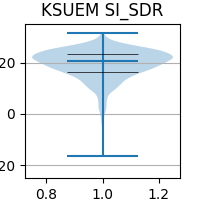}
\includegraphics[width=0.4\linewidth]{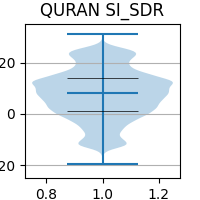}
\includegraphics[width=0.4\linewidth]{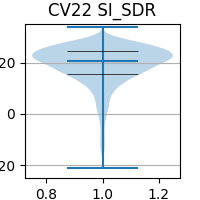}
\includegraphics[width=0.4\linewidth]{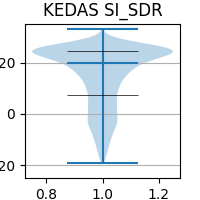}
\includegraphics[width=0.4\linewidth]{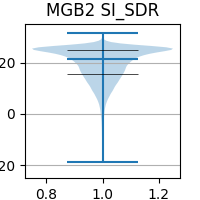}
\includegraphics[width=0.4\linewidth]{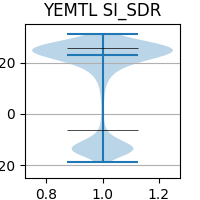}
\includegraphics[width=0.4\linewidth]{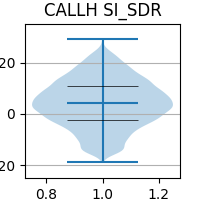}
\includegraphics[width=0.4\linewidth]{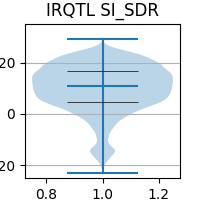}
\includegraphics[width=0.4\linewidth]{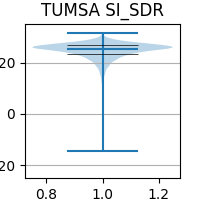}
\includegraphics[width=0.4\linewidth]{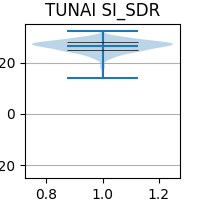}

\subsection*{SQUIM STOI}
\includegraphics[width=0.4\linewidth]{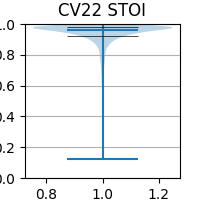}
\includegraphics[width=0.4\linewidth]{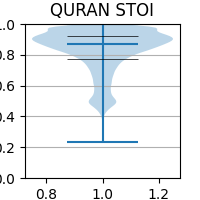}
\includegraphics[width=0.4\linewidth]{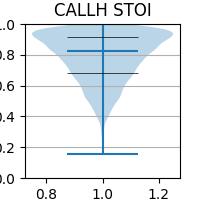}
\includegraphics[width=0.4\linewidth]{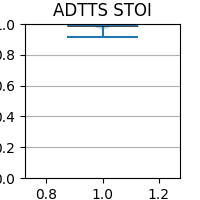}
\includegraphics[width=0.4\linewidth]{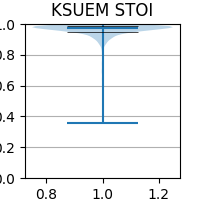}
\includegraphics[width=0.4\linewidth]{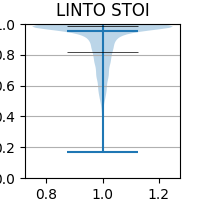}
\includegraphics[width=0.4\linewidth]{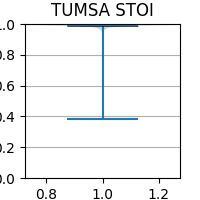}
\includegraphics[width=0.4\linewidth]{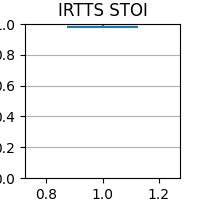}
\includegraphics[width=0.4\linewidth]{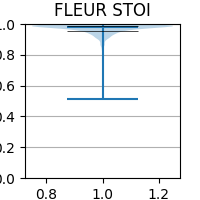}
\includegraphics[width=0.4\linewidth]{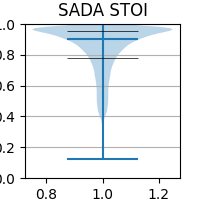}
\includegraphics[width=0.4\linewidth]{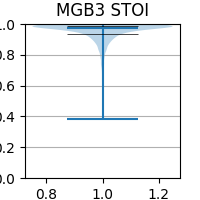}
\includegraphics[width=0.4\linewidth]{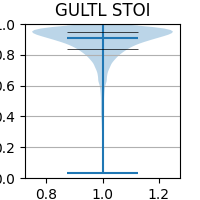}
\includegraphics[width=0.4\linewidth]{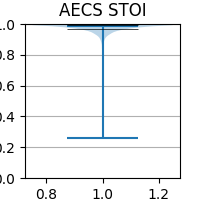}
\includegraphics[width=0.4\linewidth]{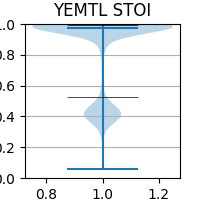}
\includegraphics[width=0.4\linewidth]{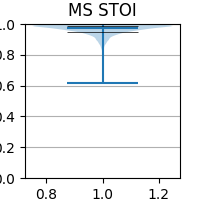}
\includegraphics[width=0.4\linewidth]{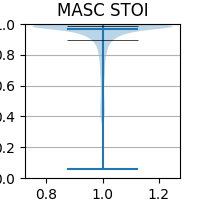}
\includegraphics[width=0.4\linewidth]{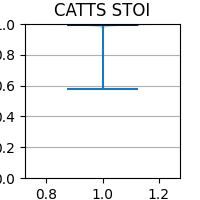}
\includegraphics[width=0.4\linewidth]{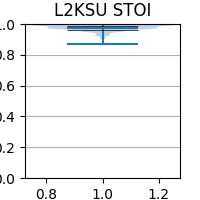}
\includegraphics[width=0.4\linewidth]{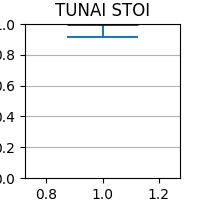}
\includegraphics[width=0.4\linewidth]{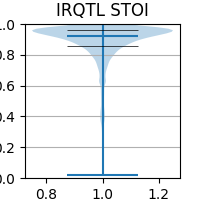}
\includegraphics[width=0.4\linewidth]{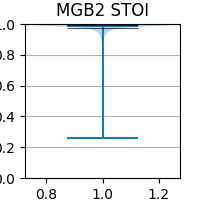}
\includegraphics[width=0.4\linewidth]{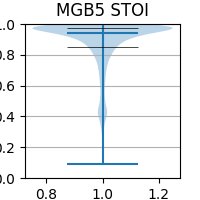}
\includegraphics[width=0.4\linewidth]{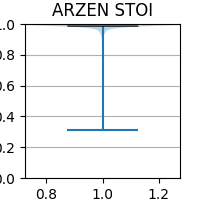}
\includegraphics[width=0.4\linewidth]{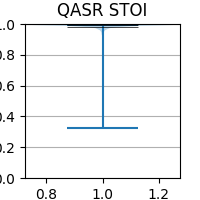}
\includegraphics[width=0.4\linewidth]{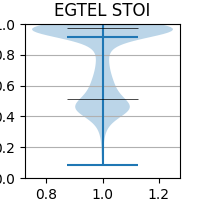}
\includegraphics[width=0.4\linewidth]{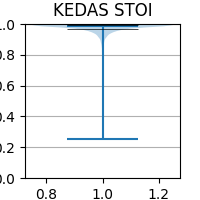}
\includegraphics[width=0.4\linewidth]{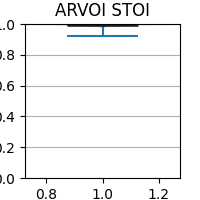}

\subsection{SQUIM NMR-MOS}
\includegraphics[width=0.4\linewidth]{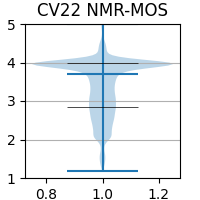}
\includegraphics[width=0.4\linewidth]{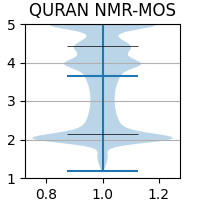}
\includegraphics[width=0.4\linewidth]{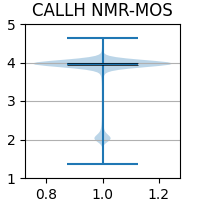}
\includegraphics[width=0.4\linewidth]{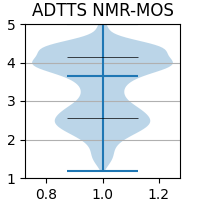}
\includegraphics[width=0.4\linewidth]{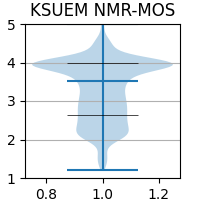}
\includegraphics[width=0.4\linewidth]{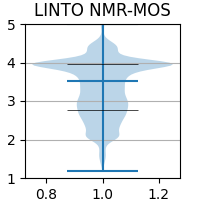}
\includegraphics[width=0.4\linewidth]{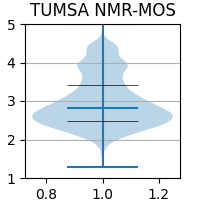}
\includegraphics[width=0.4\linewidth]{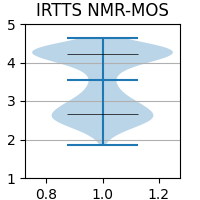}
\includegraphics[width=0.4\linewidth]{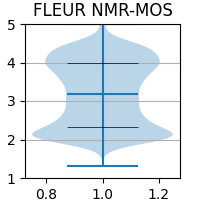}
\includegraphics[width=0.4\linewidth]{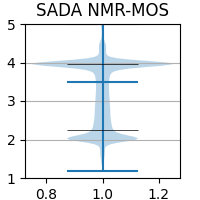}
\includegraphics[width=0.4\linewidth]{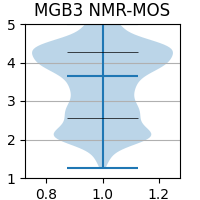}
\includegraphics[width=0.4\linewidth]{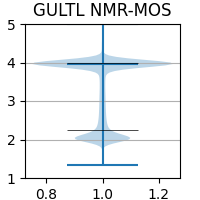}
\includegraphics[width=0.4\linewidth]{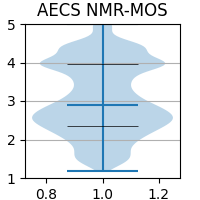}
\includegraphics[width=0.4\linewidth]{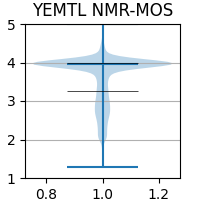}
\includegraphics[width=0.4\linewidth]{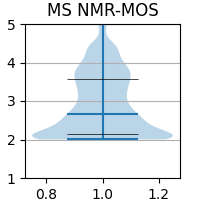}
\includegraphics[width=0.4\linewidth]{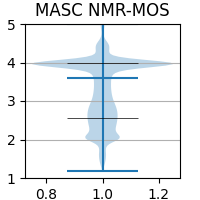}
\includegraphics[width=0.4\linewidth]{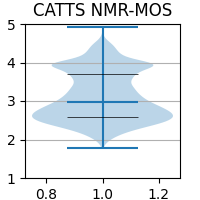}
\includegraphics[width=0.4\linewidth]{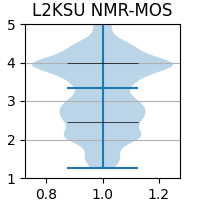}
\includegraphics[width=0.4\linewidth]{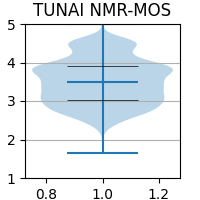}
\includegraphics[width=0.4\linewidth]{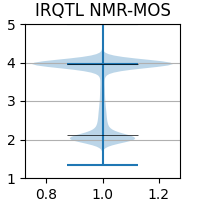}
\includegraphics[width=0.4\linewidth]{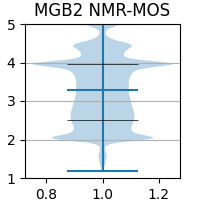}
\includegraphics[width=0.4\linewidth]{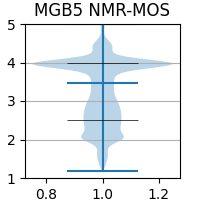}
\includegraphics[width=0.4\linewidth]{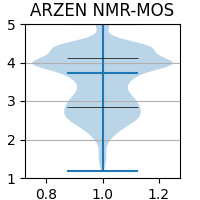}
\includegraphics[width=0.4\linewidth]{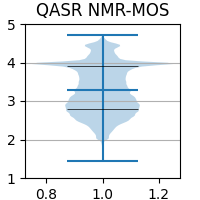}
\includegraphics[width=0.4\linewidth]{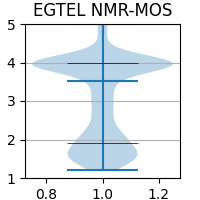}
\includegraphics[width=0.4\linewidth]{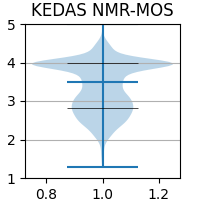}
\includegraphics[width=0.4\linewidth]{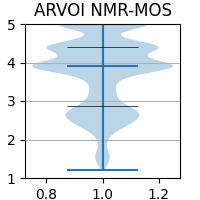}

\section{Experimental Setup} \label{appdx_sec:evaluation}

\section{Models}

\paragraph{Whisper-large-v3}\cite{radford2023robust}: This transformer-based ASR model was trained to perform long-form transcription, speech-to-text translation, and language ID using large quantities of multilingual audio sourced from internet videos with subtitles. There may be a bias towards producing MSA-like sentences, due to the decoder's language modeling capabilities combined with prevalence of writing subtitles in MSA.

\paragraph{MMS}\cite{pratap2024scaling}: In contrast to Whisper's approach, MMS uses a CTC-based model trained primarily on translated religious texts. Like Whisper, however, MMS treats all Arabic dialects as belonging to one language, however, this is less likely to be problematic in converting sentences into MSA structure due to CTC not relying on language modeling.

\paragraph{Seamless M4T v2}\cite{barrault2023seamless}: While originally designed for speech-to-speech translation, this model performs ASR by leveraging only speech encoder and text decoder of the pipeline to generate target text. Aside from MSA, this model supports two other Arabic dialects: Moroccan (ary) and Egyptian (arz). 

\paragraph{Omnilingual}\cite{omnilingual2025omnilingual}: Like many of the other models, this is also a transformer-encoder based model, with options for either CTC text decoding or transformer decoder based language modeling. What sets it appart from the other models is use of a new high quality dataset focusing on extremely low-resource languages alongside existing publicly available resources. While the exact list of open source datasets that Omnilingual was trained on is not available, they also provide a newly collected set of recordings for many Arabic dialects. In total 18 dialects are provided in this new dataset, however, of these only 13 have more than 1000 utterances across all splits including: Baharna (abv), Hijazi (acw), Omani (acx), Tunisian (aeb), Saidi (aec), Gulf (afb), Sudanese (apd), Algerian  (arq), Najdi (ars), Moroccan  (ary), Egyptian (arz), Libyan  (ayl), and North Mesopotamian (ayp).

\subsection{Normalization}\label{appd_subsec:normalization}
The normalization pipeline $\mathcal{N}(x)$ transforms an input string $x$ through a sequential cleaning process. If the text is detected as Buckwalter transliteration, it is mapped to the Arabic script. The string then undergoes Unicode NFKD decomposition, facilitating the removal of all combining characters and Arabic diacritics (\textit{Tashkeel}). Punctuation marks from both Latin and Arabic scripts---including symbols such as \AR{،}, \AR{؛}, and \AR{؟}---are eliminated. Orthographic unification is then applied: all Hamzated Alef forms \{\AR{أ}, \AR{إ}, \AR{آ}\} are normalized to bare Alef (\AR{ا}), Ta-Marbuta (\AR{ة}) is converted to Ha (\AR{ه}), and Alif-Maqsura (\AR{ى}) is standardized to Ya (\AR{ي}). Finally, all text is trimmed and consecutive whitespace is collapsed into single spaces.


\begin{table*}[h]
\centering
\renewcommand\cellalign{lc}
\resizebox{0.8\textwidth}{!}{%
\begin{tabular}{lHHcclccl}
\toprule
\multirow{2}{*}{\textbf{Variety}} & \multicolumn{2}{H}{\textbf{Train}}  & & \multicolumn{2}{c}{\textbf{Dev}} &  & \multicolumn{2}{c}{\textbf{Test}}  \\  \cmidrule{5-6}  \cmidrule{8-9}

 &\textbf{ Dur (h) }& \textbf{Dataset} &&  \textbf{ Dur (h) }& \textbf{Dataset} & & \textbf{ Dur (h) }& \textbf{Dataset}  \\

\toprule
acm & 5.0 & \makecell{(IRTTS 0.2, IRQTL 4.8, \\ MASC 0.2,) } & & 0.4 & \makecell{(IRQTL 0.7, MASC 0.3) } & & 1.0 & \makecell{ (MASC 0.1, IRQTL 0.9,) } \\
acq & 0.8 & \makecell{(YEMTL 0.8) } & & 0.9 & \makecell{(YEMTL 0.9) } &&  0.9 & \makecell{ (YEMTL 0.9) } \\
acw & 5.0 & \makecell{(SADA 5.0) } & & 0.6 & \makecell{(SADA 0.6) } & & 1.0 & \makecell{ (SADA 1.0) } \\
aeb & 0.5 & \makecell{(MASC 0.5) } & & 0.0 & \makecell{(MASC 0.0) } & & 1.0 & \makecell{ (TARIC 0.4, LINTO 0.6,) } \\
afb & 5.0 & \makecell{(GULTL 5.0, MASC 0.0) } & &  1.0 & \makecell{(MASC 0.0, CASA 0.5, \\ GULTL 0.5) } & & 1.0 & \makecell{ (CASA 0.3, MASC 0.0, \\ GULTL 0.7,) } \\
apc & 5.1 & \makecell{(TUNAI 0.6, MASC 4.3, \\ SADA 0.2,) } & & 1.0 & \makecell{(MASC 0.7, IWSLT 0.0, \\ SADA 0.0, CASA 0.3,) } & & 0.9 & \makecell{ (TUNAI 0.0, MASC 0.6, \\ IWSLT 0.0, CASA 0.3,) } \\
arb & 5.0 & \makecell{(MASC 5.0) } & & 1.0 & \makecell{(TUMSA 0.0, MASC 0.9, \\ FLEUR 0.1, SADA 0.0,) } & & 1.0 & \makecell{ (TUMSA 0.0, MASC 0.8, \\ FLEUR 0.1, SADA 0.0, \\ ARVOI 0.1,) } \\
arq & 5.0 & \makecell{(MASC 5.0) } & & 1.0 & \makecell{(CASA 1.0) } & & 1.0 & \makecell{ (MASC 0.1, CASA 0.9,) } \\
ars & 5.0 & \makecell{(SADA 5.0) } &&  1.0 & \makecell{(SADA 1.0) } & & 1.0 & \makecell{ (SADA 1.0) } \\
ary & 5.0 & \makecell{(MGB5 4.9, MASC 0.1,) } & & 1.0 & \makecell{(MGB5 0.9, MASC 0.0, \\ CASA 0.1,) } & & 1.0 & \makecell{ (MGB5 0.9, MASC 0.0, \\ CASA 0.1,) } \\
arz & 5.0 & \makecell{(MGB3 1.9, MASC 1.6, \\ SADA 0.3, ARZEN 1.0, \\ EGTEL 0.3,) } & & 1.0 & \makecell{(MGB3 0.5, MASC 0.3, \\ SADA 0.0, CASA 0.1,) } & & 1.0 & \makecell{ (MGB3 0.6, MASC 0.2, \\ SADA 0.0, CASA 0.1,) } \\
ayn & 0.7 & \makecell{(YEMTL 0.7) } & &  0.7 & \makecell{(YEMTL 0.7) } & & 0.7 & \makecell{ (YEMTL 0.7) } \\ \midrule

\textbf{Total} &\textbf{47.2} &  & &  \textbf{9.6} &  & & \textbf{11.5} &  \\ \bottomrule

\end{tabular}%
}
\caption{Benchmark Overview. Time in hours. Subsets with 0 hours indicate < 5 minutes of duration}
\label{tab:benchmark}
\end{table*}

\section{Results}\label{appdx_sec:results}
\begin{figure*}[!ht]
    \centering
    \includegraphics[width=0.98\linewidth]{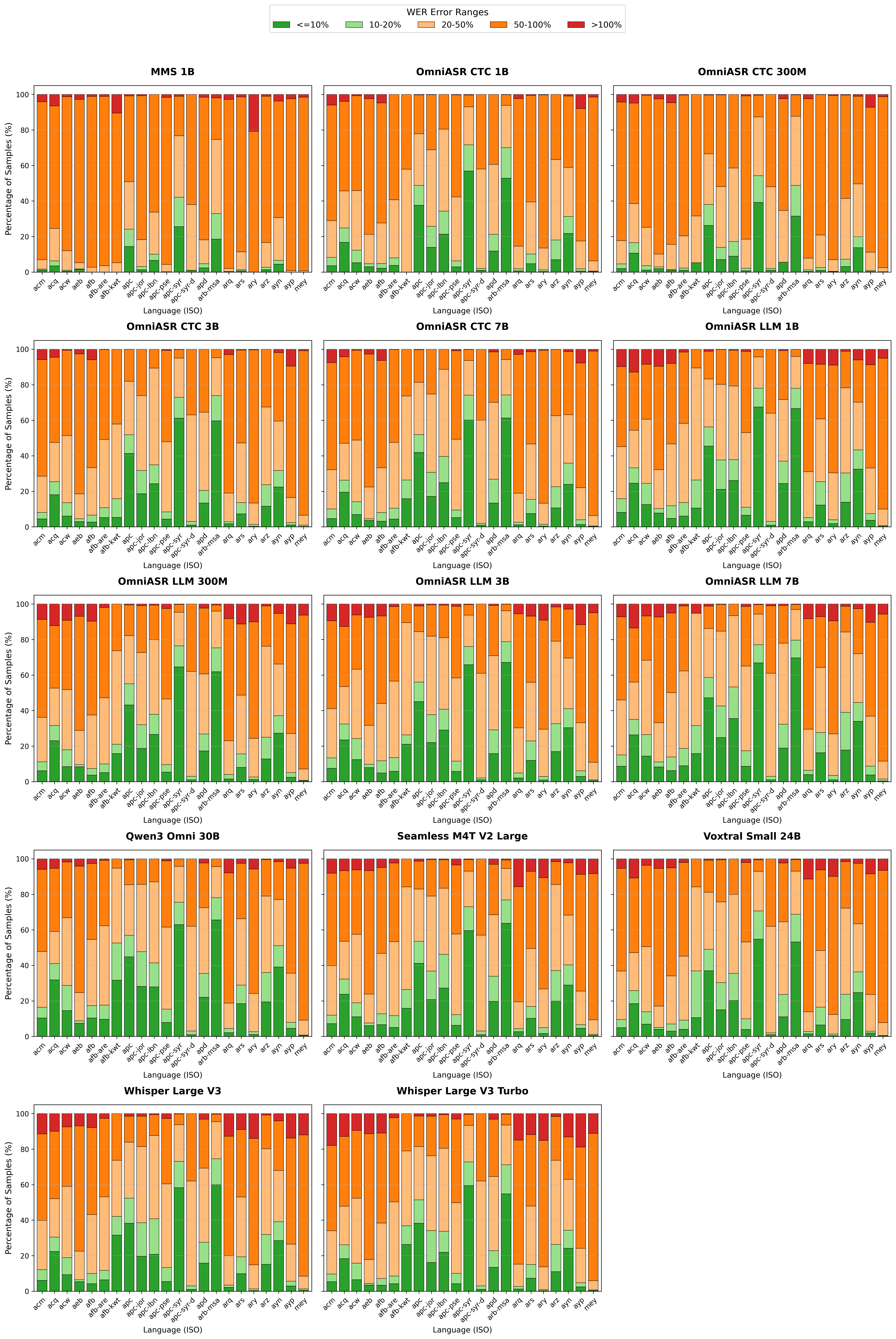}
    \caption{WER distribution across languages for all ASR models from the three architectures listed in Table~\ref{tab:asr_results_wer}. Stacked bar charts show the proportion of samples in different WER ranges (from $\leq$10\% to $>$100\%) across multiple languages (ISO codes). Colors range from green (low error rates) to red (high error rates).}
    \label{appdx_fig:WER_analysis}
\end{figure*}

\begin{figure*}[!ht]
    \centering
    \includegraphics[width=0.98\linewidth]{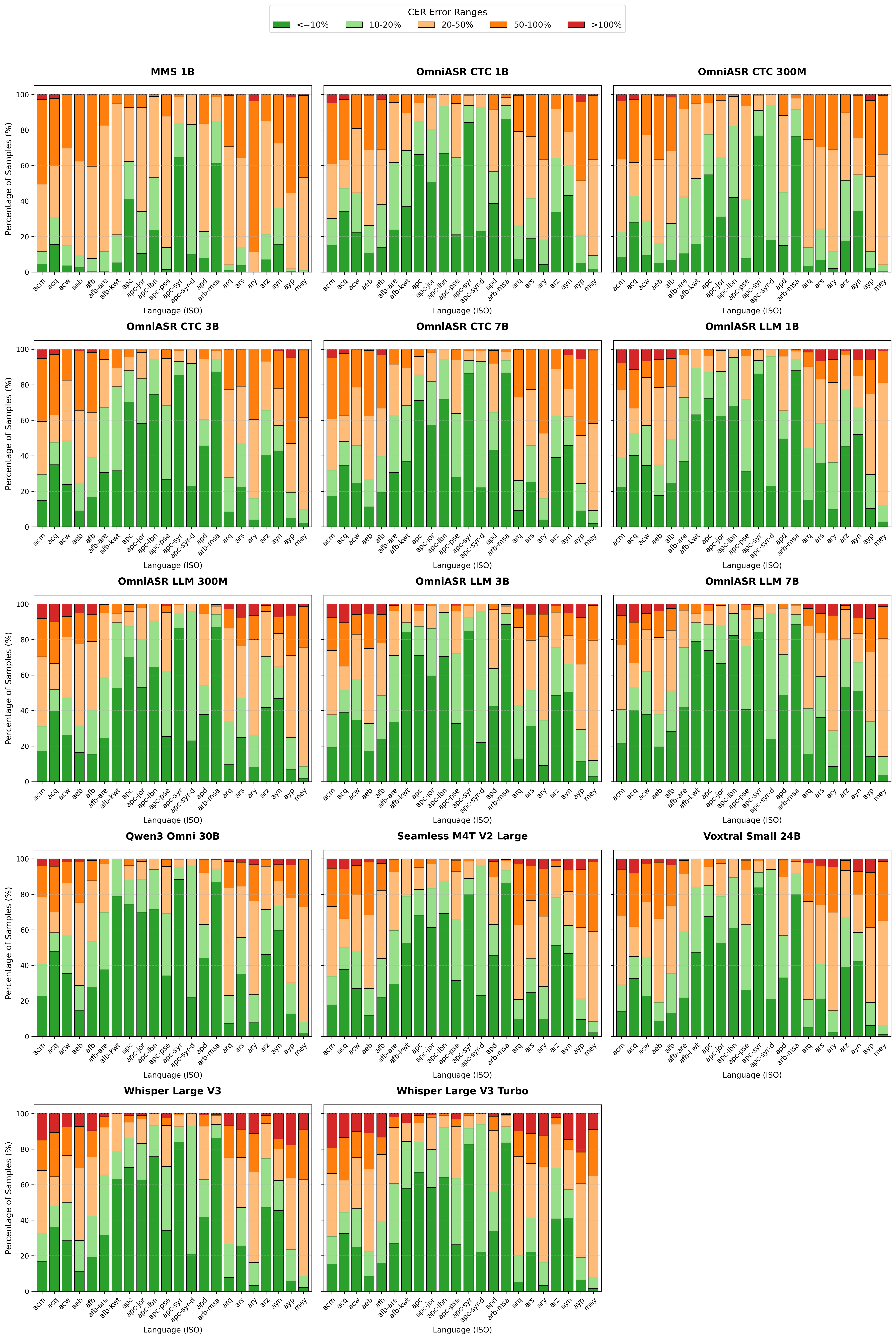}
    \caption{CER distribution across languages for all ASR models from the three architectures listed in Table~\ref{tab:asr_results_wer}. Stacked bar charts show the proportion of samples in different CER ranges (from $\leq$10\% to $>$100\%) across multiple languages (ISO codes). Colors range from green (low error rates) to red (high error rates).}
    \label{appdx_fig:CER_analysis}
\end{figure*}

Please see Tables \ref{tab:appendix_asr_results_wer} \& \ref{appdx_tab:results_cer} and Figs. \ref{appdx_fig:CER_analysis} \& \ref{appdx_fig:WER_analysis} for detailed breakdown of the performance of off the shelf models.
\begin{table*}[]
\resizebox{\textwidth}{!}{%
\begin{tabular}{lrrrrrlrrrlrrrrlr}
\toprule
\multirow{2}{*}{\textbf{Dialect}}                              & \multicolumn{5}{c}{\textbf{Encode-Only (CTC / Acoustic Encoder)}}                                                                                                          &  & \multicolumn{3}{c}{\textbf{Encode-Decoder}}                                                             &  & \multicolumn{6}{c}{\textbf{Multimodal (Speech + LLM Core)}}                                                                                                                                                     \\ \cmidrule{2-6} \cmidrule{8-10} \cmidrule{12-17}

& \multicolumn{1}{c}{\colorbox{blue!15}{\textbf{E1}}} & \multicolumn{1}{c}{\colorbox{blue!15}{\textbf{E2}}} & \multicolumn{1}{c}{\colorbox{blue!15}{\textbf{E3}}} & \multicolumn{1}{c}{\colorbox{blue!15}{\textbf{E4}}} & \multicolumn{1}{c}{\colorbox{blue!15}{\textbf{E5}}} & & \multicolumn{1}{c}{\colorbox{orange!15}{\textbf{ED1}}} & \multicolumn{1}{c}{\colorbox{orange!15}{\textbf{ED2}}} & \multicolumn{1}{c}{\colorbox{orange!15}{\textbf{ED3}}} & & \multicolumn{1}{c}{\colorbox{green!15}{\textbf{M1}}} & \multicolumn{1}{c}{\colorbox{green!15}{\textbf{M2}}} & \multicolumn{1}{c}{\colorbox{green!15}{\textbf{M3}}} & \multicolumn{1}{c}{\colorbox{green!15}{\textbf{M4}}} & \multicolumn{1}{c}{\colorbox{green!15}{\textbf{M5}}} & \multicolumn{1}{c}{\colorbox{green!15}{\textbf{M6}}} \\
\toprule

MSA (arb) & 38.94 & 26.18 & 16.49 & 13.5 & 13.8 &   & 18.51 & 15.55 & 14.7 &   & 17.52 & 11.51 & 13.57 & 11.52 & 11.3 & \textbf{10.19}\\ \midrule

Khaleeji (afb)\textsuperscript{$\star$} & 89.62 & 81.54 & 74.96 & 73.54 & 71.72 &   & 213.89 & 105.14 & 61.89 &   & 80.23 & \textbf{53.71} & 121.57 & 84.05 & 91.34 & 68.39\\
Khaleeji (Kuwait) (afb-kwt) & 86.47 & 59.6 & 58.43 & 48.37 & 45.81 &   & 35.17 & 30.85 & 33.65 &   & 34.53 & \textbf{19.45} & 39.59 & 30.92 & 30.71 & 29.48\\
Khaleeji (UAE) (afb-are) & 84.27 & 70.13 & 58.37 & 53.92 & 56.07 &   & 68.09 & 59.48 & 54.7 &   & 60.69 & 46.22 & 55.53 & 50.23 & 50.08 & \textbf{45.71}\\
Najdi (Saudi Arabia) (ars) & 84.01 & 76.93 & 64.06 & 59.31 & 60.17 &   & 173.42 & 113.92 & 62.63 &   & 72.73 & \textbf{46.9} & 80.75 & 63.92 & 61.31 & 60.23\\
Hijazi (Saudi Arabia) (acw) & 82.03 & 72.77 & 59.08 & 56.73 & 58.05 &   & 191.21 & 105.94 & 62.14 &   & 91.53 & 55.93 & 67.79 & 73.33 & 56.43 & \textbf{53.49}\\
Sanaani Arabic (Yemen) (ayn) & 74.99 & 58.25 & 50.36 & 50.49 & 47.81 &   & 188.3 & 59.06 & 44.25 &   & 47.67 & \textbf{34.63} & 52.32 & 53.7 & 46.19 & 40.37\\
Ta'izzi-Adeni Arabic (Yemen) (acq) & 72.61 & 55.22 & 46.42 & 45.84 & 45.44 &   & 125.22 & 77.06 & 39.54 &   & 50.62 & \textbf{28.77} & 44.02 & 43.43 & 38.03 & 35.5\\ \midrule
North Mesopotamian Arabic (Iraq) (ayp) & 94.65 & 89.14 & 87.58 & 87.99 & 84.11 &   & 367.74 & 134.53 & 87.78 &   & 99.13 & \textbf{75.45} & 88.48 & 93.12 & 104.72 & 95.16\\ 
Mesopotamian Arabic (Iraq) (acm) & 94.02 & 86.43 & 80.36 & 81.2 & 80.13 &   & 271.25 & 119.91 & 106.26 &   & 87.53 & \textbf{71.16} & 108.36 & 86.76 & 81.93 & 73.86\\ \midrule
Levantine (apc)\textsuperscript{$\star$} & 51.75 & 38.91 & 29.72 & 26.84 & 26.88 &   & 36.31 & 32.53 & 26.38 &   & 28.28 & 23.81 & 27.61 & 25.45 & 25.17 & \textbf{23.53}\\
Levantine (Jordan) (apc-jor) & 72.89 & 54.07 & 40.51 & 36.08 & 36.61 &   & 46.78 & 41.26 & 33.21 &   & 37.46 & \textbf{27.2} & 37.29 & 32.37 & 32.53 & 29.31\\
Levantine (Lebanon) (apc-lbn) & 62.63 & 46.49 & 32.61 & 29.68 & 28.33 &   & 32.58 & 29.19 & 29.14 &   & 32.05 & 27.1 & 32.58 & 32.18 & 30.12 & \textbf{21.56}\\
Levantine (Palestine) (apc-pse) & 82.74 & 70.23 & 57.58 & 54.74 & 54.28 &   & 73.17 & 53.82 & 50.61 &   & 55.29 & 47.2 & 55.58 & 51.7 & 50.53 & \textbf{45.63}\\
Levantine (Syria) (apc-syr) & 34.17 & 23.64 & 15.11 & 13.59 & 13.81 &   & 19.15 & 14.33 & 14.54 &   & 16.34 & 12.02 & 12.28 & \textbf{11.23} & 12.43 & 12.3\\
Levantine (Syrian Damascus) (apc-syr-d) & 55.97 & 52.04 & 49.36 & 48.63 & 49.16 &   & 48.65 & 48.74 & 49.34 &   & 49.15 & 48.69 & 48.68 & \textbf{47.87} & 49.1 & 49.03\\ \midrule
Egyptian (arz) & 74.32 & 58.87 & 45.7 & 42.28 & 45.59 &   & 48.16 & 35.69 & 37.61 &   & 43.25 & 68.6 & 41.7 & 49.38 & 48.76 & \textbf{34.8}\\
Sudanese (apd) & 74.82 & 63.97 & 49.1 & 45.71 & 44.12 &   & 70.9 & 49.19 & 47.26 &   & 48.71 & 40.06 & 45.49 & 37.68 & 41.33 & \textbf{36.01}\\ \midrule
Algerian (arq) & 96.19 & 88.26 & 81.65 & 78.64 & \textbf{78.01} &   & 175.34 & 116.92 & 124.2 &   & 109.76 & 82.56 & 83.6 & 79.25 & 78.92 & 86.29\\
Hassaniyya (mey) & 96.57 & 91.79 & 89.19 & 88.55 & 88.45 &   & 204.97 & 135.94 & 93.46 &   & 106.15 & 83.52 & 88.24 & 84.3 & 84.58 & \textbf{83.17}\\
Moroccan (ary) & 113.83 & 84.48 & 79.78 & \textbf{79.71} & 81.77 &   & 164.21 & 120.43 & 86.2 &   & 122.79 & 99.33 & 94.68 & 98.85 & 93.19 & 103.8\\
Tunisian (aeb) & 92.46 & 86.38 & 78.37 & 80.51 & 79.33 &   & 225.04 & 111.93 & 80.63 &   & 85.32 & 247.39 & 80.62 & 84.17 & 77.55 & \textbf{74.31}\\
\bottomrule
\end{tabular}%
}
\caption{A comprehensive comparison of WER performance across Arabic dialect varieties for 14 models spanning three distinct ASR architectural paradigms on  TEST dataset. \textbf{Encode-Only:} \colorbox{blue!15}{\textbf{E1.}} MMS-1B-ALL, \colorbox{blue!15}{\textbf{E2.}} omniASR-CTC-300M-v2, \colorbox{blue!15}{\textbf{E3.}} omniASR-CTC-1B-v2, \colorbox{blue!15}{\textbf{E4.}} omniASR-CTC-3B-v2, and \colorbox{blue!15}{\textbf{E5.}} omniASR-CTC-7B-v2. \textbf{Encode-Decoder:} \colorbox{orange!15}{\textbf{ED1.}} Whisper-Large-v3-Turbo, \colorbox{orange!15}{\textbf{ED2.}} Whisper-Large-v3, and \colorbox{orange!15}{\textbf{ED3.}} SeamlessM4T-v2-Large. \textbf{Multimodal:} \colorbox{green!15}{\textbf{M1.}} Voxtral-Small-24B-2507, \colorbox{green!15}{\textbf{M2.}} Qwen3-Omni-30B-A3B-Instruct, \colorbox{green!15}{\textbf{M3.}} omniASR-LLM-300M-v2, \colorbox{green!15}{\textbf{M4.}} omniASR-LLM-1B-v2, \colorbox{green!15}{\textbf{M5.}} omniASR-LLM-3B-v2, and \colorbox{green!15}{\textbf{M6.}} omniASR-LLM-7B-v2. \textbf{Bold} refers to the best performance for each dialect. \textsuperscript{$\star$}Khaleeji (afb) and Levantine (apc) include varieties from multiple countries. \textit{For CER performance, details are provided in Table~\ref{appdx_tab:results_cer} (\S~\ref{appdx_sec:results}). }}
\label{tab:appendix_asr_results_wer}

\end{table*}
\begin{table*}[!ht]
\resizebox{\textwidth}{!}{%
\begin{tabular}{lrrrrrlrrrlrrrrlr}
\toprule
\multirow{2}{*}{\textbf{Dialect}}                              & \multicolumn{5}{c}{\textbf{Encode-Only (CTC / Acoustic Encoder)}}                                                                                                          &  & \multicolumn{3}{c}{\textbf{Encode-Decoder}}                                                             &  & \multicolumn{6}{c}{\textbf{Multimodal (Speech + LLM Core)}}                                                                                                                                                     \\ \cmidrule{2-6} \cmidrule{8-10} \cmidrule{12-17}

& \multicolumn{1}{c}{\colorbox{blue!15}{\textbf{E1}}} & \multicolumn{1}{c}{\colorbox{blue!15}{\textbf{E2}}} & \multicolumn{1}{c}{\colorbox{blue!15}{\textbf{E3}}} & \multicolumn{1}{c}{\colorbox{blue!15}{\textbf{E4}}} & \multicolumn{1}{c}{\colorbox{blue!15}{\textbf{E5}}} & & \multicolumn{1}{c}{\colorbox{orange!15}{\textbf{ED1}}} & \multicolumn{1}{c}{\colorbox{orange!15}{\textbf{ED2}}} & \multicolumn{1}{c}{\colorbox{orange!15}{\textbf{ED3}}} & & \multicolumn{1}{c}{\colorbox{green!15}{\textbf{M1}}} & \multicolumn{1}{c}{\colorbox{green!15}{\textbf{M2}}} & \multicolumn{1}{c}{\colorbox{green!15}{\textbf{M3}}} & \multicolumn{1}{c}{\colorbox{green!15}{\textbf{M4}}} & \multicolumn{1}{c}{\colorbox{green!15}{\textbf{M5}}} & \multicolumn{1}{c}{\colorbox{green!15}{\textbf{M6}}} \\
\toprule

MSA (arb) & 11.12 & 8.24 & 5.64 & 4.66 & 5.11 &   & 12.35 & 9.38 & 7.04 &   & 6.64 & \textbf{4.18} & 5.26 & 4.43 & 4.29 & 4.2\\ \midrule
Khaleeji (afb)\textsuperscript{$\star$} & 49.85 & 45.03 & 42.93 & 43.3 & 40.94 &   & 344.06 & 176.63 & \textbf{33.42} &   & 59.15 & 63.7 & 134.19 & 59.73 & 80.09 & 46.23\\
Khaleeji (Kuwait) (afb-kwt) & 27.97 & 21.45 & 20.34 & 18.39 & 17.55 &   & 118.03 & 11.74 & 11.91 &   & 13.95 & \textbf{4.79} & 13.09 & 11.17 & 8.51 & 8.77\\
Khaleeji (UAE) (afb-are) & 37 & 26.76 & 20.76 & 19.3 & 21.45 &   & 40.36 & 33.09 & 22.6 &   & 28.34 & 16.6 & 21.3 & 17.56 & 20.41 & \textbf{15.7}\\
Hijazi (Saudi Arabia) (acw) & 43.13 & 37.85 & 31.29 & \textbf{29.67} & 32.27 &   & 203.55 & 133.56 & 41.81 &   & 61.28 & 34.98 & 43.87 & 52.56 & 44.51 & 35.43\\
Najdi (Saudi Arabia) (ars) & 46.25 & 42.98 & 35.81 & 32.51 & 33.59 &   & 213.69 & 148.74 & 35.73 &   & 63.07 & \textbf{26.73} & 54.22 & 44.12 & 38.91 & 44.2\\
Sanaani Arabic (Yemen) (ayn) & 40.01 & 33.54 & 28.84 & 30.73 & 31.77 &   & 421.29 & 1071.03 & 32.35 &   & 584.01 & \textbf{20.56} & 79.7 & 55.51 & 50.19 & 96.5\\
Ta'izzi-Adeni Arabic (Yemen) (acq) & 31.87 & 25.71 & \textbf{22.2} & 22.4 & 23.26 &   & 185.68 & 216.12 & 24.08 &   & 64.58 & 32.23 & 50.56 & 50.34 & 41.98 & 80.95\\ \midrule
Mesopotamian Arabic (Iraq) (acm) & 67.35 & 65.13 & \textbf{62.33} & 66.98 & 64.05 &   & 479.24 & 230.88 & 100.67 &   & 69.87 & 70.19 & 180.22 & 96.86 & 83.67 & 136.4\\
North Mesopotamian Arabic (Iraq) (ayp) & 60.59 & 60.03 & 58.99 & 61.9 & 57.16 &   & 546.65 & 296.25 & 61.76 &   & 113.44 & \textbf{53.94} & 81.48 & 105.31 & 94.37 & 104.4\\ \midrule
Levantine (apc)\textsuperscript{$\star$} & 19.14 & 14.28 & 11.27 & 10.19 & 10.28 &   & 21.26 & 17.07 & 13 &   & 10.88 & \textbf{8.99} & 11.95 & 9.78 & 9.9 & 9.3\\
Levantine (Jordan) (apc-jor) & 27.66 & 18.53 & 13.16 & 11.75 & 12.1 &   & 23.96 & 21.41 & 11.79 &   & 13.69 & \textbf{8.89} & 12.44 & 9.88 & 10.17 & 9.26\\
Levantine (Lebanon) (apc-lbn) & 19.69 & 13.24 & 8.42 & 7.31 & 7.43 &   & 9.2 & 7.57 & 9.82 &   & 9.48 & 7.34 & 8.56 & 7.76 & 7.55 & \textbf{5.69}\\
Levantine (Palestine) (apc-pse) & 34.26 & 26.45 & 20.89 & 19.55 & 20.3 &   & 42.53 & 32.05 & 19.93 &   & 22.85 & 17.86 & 25.32 & 18.09 & 18.61 & \textbf{15.81}\\
Levantine (Syria) (apc-syr) & 10.16 & 7.1 & 4.95 & 4.86 & 4.75 &   & 11.05 & 5.12 & 6.18 &   & 5.97 & \textbf{3.82} & 4.08 & 4.38 & 4.65 & 5.38\\
Levantine (Syrian Damascus) (apc-syr-d) & 15.77 & 14.11 & 13.93 & 13.93 & 13.89 &   & 13.54 & 13.77 & 13.35 &   & 13.71 & 13.48 & \textbf{13.17} & 13.24 & 13.43 & 13.57\\ \midrule
Egyptian (arz) & 33.9 & 25.43 & 20.58 & \textbf{19.38} & 22.42 &   & 46.96 & 31.61 & 23.88 &   & 23.28 & 43.07 & 30.1 & 43.96 & 40.32 & 27.72\\
Sudanese (apd) & 34.23 & 27.99 & 21.45 & 19.18 & 19.87 &   & 51.6 & 25.05 & 24.41 &   & 23.69 & 1630.3 & 20.18 & 15.17 & 17.76 & \textbf{14.94}\\ \midrule
Algerian (arq) & 48.5 & 43.51 & 38.99 & \textbf{37.22} & 39.16 &   & 139.05 & 112.26 & 69.75 &   & 79.5 & 632.82 & 51.55 & 37.43 & 45.21 & 44.96\\
Hassaniyya (mey) & 54.26 & 49.52 & 49.79 & 50.2 & 51.84 &   & 161.86 & 109.66 & 54.03 &   & 71.11 & 44.39 & 46.84 & \textbf{42.21} & 42.46 & 43.85\\
Tunisian (aeb) & 46.91 & 46.03 & 40.49 & 42.99 & 43.8 &   & 246.38 & 150.85 & 45.25 &   & 47.16 & 1244.96 & 48.03 & 52.23 & 47.3 & \textbf{39.27}\\
Moroccan (ary) & 69.71 & \textbf{45.64} & 46.43 & 47.22 & 52.7 &   & 138.48 & 85.35 & 51.93 &   & 87.56 & 63.29 & 60.89 & 66.58 & 61.64 & 85.6\\
\bottomrule
\end{tabular}%
}
\caption{A comprehensive comparison of CER performance across Arabic dialect varieties for 14 models spanning three distinct ASR architectural paradigms on  TEST dataset. \textbf{Encode-Only:} \colorbox{blue!15}{\textbf{E1.}} MMS-1B-ALL, \colorbox{blue!15}{\textbf{E2.}} omniASR-CTC-300M-v2, \colorbox{blue!15}{\textbf{E3.}} omniASR-CTC-1B-v2, \colorbox{blue!15}{\textbf{E4.}} omniASR-CTC-3B-v2, and \colorbox{blue!15}{\textbf{E5.}} omniASR-CTC-7B-v2. \textbf{Encode-Decoder:} \colorbox{orange!15}{\textbf{ED1.}} Whisper-Large-v3-Turbo, \colorbox{orange!15}{\textbf{ED2.}} Whisper-Large-v3, and \colorbox{orange!15}{\textbf{ED3.}} SeamlessM4T-v2-Large. \textbf{Multimodal:} \colorbox{green!15}{\textbf{M1.}} Voxtral-Small-24B-2507, \colorbox{green!15}{\textbf{M2.}} Qwen3-Omni-30B-A3B-Instruct, \colorbox{green!15}{\textbf{M3.}} omniASR-LLM-300M-v2, \colorbox{green!15}{\textbf{M4.}} omniASR-LLM-1B-v2, \colorbox{green!15}{\textbf{M5.}} omniASR-LLM-3B-v2, and \colorbox{green!15}{\textbf{M6.}} omniASR-LLM-7B-v2. \textbf{Bold} refers to the best performance for each dialect. \textsuperscript{$\star$}Khaleeji (afb) and Levantine (apc) include varieties from multiple countries. } \label{appdx_tab:results_cer}

\end{table*}

\begin{table*}[!ht]
\resizebox{\textwidth}{!}{%
\begin{tabular}{lrrrrrlrrrlrrrrlr}
\toprule
\multirow{2}{*}{\textbf{Dialect}}                              & \multicolumn{5}{c}{\textbf{Encode-Only (CTC / Acoustic Encoder)}}                                                                                                          &  & \multicolumn{3}{c}{\textbf{Encode-Decoder}}                                                             &  & \multicolumn{6}{c}{\textbf{Multimodal (Speech + LLM Core)}}                                                                                                                                                     \\ \cmidrule{2-6} \cmidrule{8-10} \cmidrule{12-17}

& \multicolumn{1}{c}{\colorbox{blue!15}{\textbf{E1}}} & \multicolumn{1}{c}{\colorbox{blue!15}{\textbf{E2}}} & \multicolumn{1}{c}{\colorbox{blue!15}{\textbf{E3}}} & \multicolumn{1}{c}{\colorbox{blue!15}{\textbf{E4}}} & \multicolumn{1}{c}{\colorbox{blue!15}{\textbf{E5}}} & & \multicolumn{1}{c}{\colorbox{orange!15}{\textbf{ED1}}} & \multicolumn{1}{c}{\colorbox{orange!15}{\textbf{ED2}}} & \multicolumn{1}{c}{\colorbox{orange!15}{\textbf{ED3}}} & & \multicolumn{1}{c}{\colorbox{green!15}{\textbf{M1}}} & \multicolumn{1}{c}{\colorbox{green!15}{\textbf{M2}}} & \multicolumn{1}{c}{\colorbox{green!15}{\textbf{M3}}} & \multicolumn{1}{c}{\colorbox{green!15}{\textbf{M4}}} & \multicolumn{1}{c}{\colorbox{green!15}{\textbf{M5}}} & \multicolumn{1}{c}{\colorbox{green!15}{\textbf{M6}}} \\
\toprule
MSA (arb) & 40.31 & 25.85 & 16.6 & 14.17 & 14.08 &   & 26.15 & 17.86 & 13.07 &   & 17.07 & 16.52 & 13.22 & 11.48 & 11.33 & \textbf{10.13}\\ \midrule
Khaleeji  (afb)\textsuperscript{$\star$} & 89.09 & 79.41 & 70.57 & 68.74 & 66.81 &   & 150.95 & 86.83 & 62.98 &   & 72.49 & \textbf{52.53} & 128.76 & 86.49 & 70.22 & 62.97\\ 
Khaleeji (Kuwait) (afb-kwt) & 83.55 & 74.97 & 64.79 & 56.75 & 57.39 &   & 44.33 & 42.83 & 36.55 &   & 54.32 & \textbf{33.74} & 53.4 & 37.46 & 38.79 & 34.83\\
Khaleeji (UAE) (afb-are) & 82.29 & 68.24 & 55.97 & 52.4 & 53.86 &   & 70.55 & 52.77 & 50.49 &   & 55.49 & 44.7 & 52.97 & 46.68 & 47.1 & \textbf{43.1}\\
Najdi (Saudi Arabia) (ars) & 78.63 & 72.12 & 57.95 & 55.11 & 53.78 &   & 124.95 & 90.84 & 58.51 &   & 60.37 & \textbf{40.58} & 69.34 & 65.38 & 59.65 & 46.51\\
Hijazi (Saudi Arabia) (acw) & 85.31 & 76.52 & 63.79 & 59.9 & 58.49 &   & 169.05 & 82.93 & 57.85 &   & 81.76 & \textbf{48.02} & 74.38 & 68.81 & 57 & 51.47\\
Sanaani Arabic (Yemen) (ayn) & 73.08 & 59.38 & 52.59 & 51.52 & 49.86 &   & 166.52 & 74.14 & 44.82 &   & 49.34 & \textbf{33.94} & 51.35 & 54.62 & 45.82 & 39.38\\
Ta'izzi-Adeni Arabic (Yemen) (acq) & 70.12 & 53.66 & 44.97 & 43.64 & 41.09 &   & 106.86 & 71.94 & 35.75 &   & 47.04 & \textbf{27.23} & 42.54 & 44.38 & 38.88 & 33.1\\ \midrule
North Mesopotamian Arabic (Iraq) (ayp) & 96.21 & 90.71 & 88.26 & 87.66 & 86.77 &   & 325.71 & 195.41 & 85.67 &   & 100.19 & \textbf{69.45} & 81.45 & 88.88 & 95.12 & 98.45\\ 
Mesopotamian Arabic (Iraq) (acm) & 83.85 & 75.27 & 68.37 & 67.74 & 67.46 &   & 255.35 & 139.37 & 61.58 &   & 74.09 & \textbf{55.21} & 63.68 & 77.97 & 73.88 & 58.2\\\midrule
Levantine (apc) & 55.54 & 38.09 & 28.08 & 26.14 & 26.91 &   & 29.24 & 26.53 & 34.23 &   & 28.79 & \textbf{22.46} & 25.42 & 23.71 & 23.82 & 24.16\\
Levantine (Jordan) (apc-jor) & 69.22 & 51.68 & 38.36 & 35.94 & 35.7 &   & 39.65 & 35.81 & 33.88 &   & 36 & \textbf{26.79} & 36.07 & 32.44 & 33.12 & 28.74\\
Levantine (Lebanon) (apc-lbn) & 65.13 & 42.44 & 29.95 & 26.38 & 28.43 &   & 29.78 & 25.96 & 60.9 &   & 31.31 & \textbf{21.73} & 26.2 & 23.42 & 23.21 & 33.11\\
Levantine (Palestine) (apc-pse) & 81.09 & 69.02 & 56.57 & 52.31 & 54.94 &   & 61.39 & 50.93 & 49.88 &   & 52.52 & 45.78 & 54.21 & 49.36 & 48.21 & \textbf{44.78}\\
Levantine (Syria) (apc-syr) & 35.83 & 19.27 & 11.86 & 11.52 & 10.57 &   & 13.73 & 10.56 & 10.2 &   & 12.84 & 10.2 & 8.44 & \textbf{7.59} & 8.38 & 12.5\\\midrule
Egyptian (arz) & 73.39 & 54.89 & 41.65 & 38.25 & 38.92 &   & 43.38 & 52.88 & 31.25 &   & 44.54 & 66.03 & 37 & 32.77 & 31.47 & \textbf{28.84}\\ \midrule
Algerian (arq) & 93.92 & 86.89 & 80.81 & 79.95 & 79.11 &   & 138.36 & 111.81 & 107.73 &   & 90.22 & 79.77 & 86.15 & \textbf{73.7} & 75.22 & 79.43\\
Hassaniyya (mey) & 99.79 & 94.27 & 91.66 & 91.22 & 90.62 &   & 203.92 & 165.2 & 96.01 &   & 100.48 & 88.59 & 92.21 & 90.99 & 88.77 & \textbf{86.97}\\
Moroccan (ary) & 111.94 & 88.36 & 85.3 & \textbf{85.01} & 85.96 &   & 159.77 & 136.55 & 123.77 &   & 109.74 & 88.73 & 104.17 & 108.49 & 95.71 & 107.28\\
Tunisian (aeb) & 60.19 & 53.76 & 34.93 & 27.81 & 27.77 &   & 25.33 & 19.79 & \textbf{17.35} &   & 25.54 & 20.27 & 29.48 & 21.29 & 20.77 & 23.24\\

\bottomrule
\end{tabular}%
}
\caption{A comprehensive comparison of WER performance across Arabic dialect varieties for 14 models spanning three distinct ASR architectural paradigms on the  development (DEV) dataset. \textbf{Encode-Only:} \colorbox{blue!15}{\textbf{E1.}} MMS-1B-ALL, \colorbox{blue!15}{\textbf{E2.}} omniASR-CTC-300M-v2, \colorbox{blue!15}{\textbf{E3.}} omniASR-CTC-1B-v2, \colorbox{blue!15}{\textbf{E4.}} omniASR-CTC-3B-v2, and \colorbox{blue!15}{\textbf{E5.}} omniASR-CTC-7B-v2. \textbf{Encode-Decoder:} \colorbox{orange!15}{\textbf{ED1.}} Whisper-Large-v3-Turbo, \colorbox{orange!15}{\textbf{ED2.}} Whisper-Large-v3, and \colorbox{orange!15}{\textbf{ED3.}} SeamlessM4T-v2-Large. \textbf{Multimodal:} \colorbox{green!15}{\textbf{M1.}} Voxtral-Small-24B-2507, \colorbox{green!15}{\textbf{M2.}} Qwen3-Omni-30B-A3B-Instruct, \colorbox{green!15}{\textbf{M3.}} omniASR-LLM-300M-v2, \colorbox{green!15}{\textbf{M4.}} omniASR-LLM-1B-v2, \colorbox{green!15}{\textbf{M5.}} omniASR-LLM-3B-v2, and \colorbox{green!15}{\textbf{M6.}} omniASR-LLM-7B-v2. \textbf{Bold} refers to the best performance for each dialect. \textsuperscript{$\star$}Khaleeji (afb) and Levantine (apc) include varieties from multiple countries. } \label{appdx_tab:results_wer_dev}

\end{table*}

\subsection{Human Sanity Check}\label{app_subsec:human}
To ensure the reliability of our automated profiling tools, we conducted a human-in-the-loop validation process focusing on two critical dimensions: audio quality and dialect identification. A stratified sample of the datasets was reviewed by native Arabic speakers to corroborate the output of our automated models.

\paragraph{Audio Quality Validation}
We compared our automated Perceptual Evaluation of Speech Quality scores against human Mean Opinion Scores (MOS). While the automated metrics generally correlated with human perception, our analysis revealed distinct domain-specific biases. The automated models frequently underestimated the quality of datasets containing expressive prosody or distinct acoustic environments. For instance, in the \texttt{ksuemotions} and \texttt{quran-speech} datasets, automated scoring yielded low values (approx. 1.15--1.55), whereas human annotators rated the quality as high (4.0/5.0). This indicates that standard reference-free metrics may penalize the silence intervals or dynamic range inherent to these domains. Conversely, synthesized speech (e.g., \texttt{arabic-diacritized-tts}) occasionally received high model scores despite lower human ratings, highlighting the model's insensitivity to certain unnatural robotic artifacts.

\paragraph{Dialectness Validation}
We further validated the automated dialect labels using two granularities: a binary classification (MSA vs. Dialectal Arabic) and a fine-grained dialect identification (ALDI).
\begin{itemize}
    \item \textbf{Binary Classification:} We observed near-perfect alignment between automated predictions and human annotation for the binary distinction between MSA and Dialectal Arabic, with accuracy scores approaching 100\% across most datasets. This confirms that current tools are highly robust at distinguishing standard Arabic from regional varieties.
    \item \textbf{Level of dialectness:} The ALDi model achieved an accuracy of 91\% (n=90) with 6 samples scoring low-dialectal when they should have been high, and 2 samples with segmentation errors in the transcription.
\end{itemize}
This divergence underscores the necessity of our mapping framework: while broad categories are easily automated, precise dialect mapping still benefits significantly from human verification and metadata standardization.

\end{document}